# Beyond 'One Language, One Script': Quantifying Orthographic Bias in Multilingual VLMs with PuMVR

**Prabhjot Singh**[1,2] **Bhushan Pawar**[1] **Madhu Reddiboina**[1]

[1]RediMinds Inc., USA
[2]The University of Texas at Austin, USA

{prabhjot.singh, bhushan.pawar, madhu.reddiboina}@rediminds.com

## Abstract

Current Vision-Language Models (VLMs) are celebrated for their multilingual capabilities, yet they operate under a flawed assumption: that one language corresponds to a single writing system. This overlooks billions of users of multi-script languages like Punjabi, Serbian, Hindi-Urdu, Kurdish, among many others, for whom a model's capability may be fractured by orthographic bias. We introduce **PuMVR** (Punjabi Multimodal Visual Reasoning), the first benchmark designed to quantify script-dependent bias through 375 culturally grounded image-reasoning tasks across Punjabi's three active scripts (Gurmukhi, Shahmukhi, Roman). Evaluating 10 state-of-the-art VLMs, we expose a substantial **Script Gap**: models frequently solve visual puzzles in one script while failing identical tasks in another, with accuracy deltas reaching 16% and Script Consistency Rates (SCR) as low as 24.8%. Crucially, visual input boosts absolute performance but does not close this gap, the relative bias persists. Our analysis suggests reasoning patterns show limited cross-script transferability, and Chain-of-Thought pathways diverge based on script alone. We propose SCR as a core metric for script-agnostic evaluation, challenging current multilingual assessment paradigms and providing a framework for equitable AI.

## 1 Introduction

The rapid evolution of Multimodal Vision-Language Models (VLMs) has been accompanied by claims of broad linguistic competence. Frontier systems such as GPT-4o (OpenAI, 2024) and Gemini 1.5 (Team et al., 2024) are validated on their ability to reason across dozens of languages. However, these evaluations rest on a precarious assumption: the **"One Language, One Script" (OLOS) paradigm**. By treating orthography as a deterministic function of language, current benchmarks, including XM3600 (Thapliyal et al., 2022) and MaXM (Changpinyo et al., 2023), overlook the reality of hundreds of millions of users who navigate the world through multi-script systems.

This evaluation gap reveals important questions about model explainability. If a model's reasoning fluctuates when identical semantic content is presented in different scripts, its knowledge may rely more heavily on orthographic patterns than language-level semantic concepts. Script variation serves as a critical diagnostic probe: if a visual concept understood in one script is inaccessible in another, the model's "multilinguality" may rely more on pattern matching than purely semantic grounding.

This work exposes an equity gap: benchmarks assume script-language isomorphism, which misrepresents performance for script-switching users. For a Punjabi speaker using Roman and Gurmukhi, a model that succeeds in one script but fails in the other is not partially capable, it is unreliable. We introduce Script Consistency Rate (SCR) to reframe multilingual evaluation from language breadth to orthographic robustness, establishing script-agnosticism as a requirement for true multilingual AI.

We interrogate this *Script-Reality Gap* using Punjabi as a natural laboratory. Spoken by over 125 million people, Punjabi operates through three distinct systems: Gurmukhi (Indic script), Shahmukhi (Perso-Arabic), and Roman (Latin transliteration). We introduce **PuMVR** (Punjabi Multimodal Visual Reasoning), a parallel-script benchmark of 375 expert-curated tasks isolating orthography as an independent variable in multimodal reason-ing.

Our investigation reveals that script variation fundamentally alters observable reasoning behavior. Through Chain-of-Thought (CoT) analysis, we show that script shifts cause models to attend to divergent visual features for identical image-text pairs, a phenomenon we term *reasoning divergence*. While CoT outputs are an imperfect proxy for internal computation, the high degree of orthographic instability suggests a fundamental lack of cross-modal semantic grounding.

**Our contributions are:** (1) **The PuMVR Benchmark**: 375 tasks across six categories, providing the first controlled, three-way orthographic evaluation for multimodal reasoning. (2) **Quantification of Script Bias**: A systematic audit of 10 state-of-the-art VLMs revealing significant performance gaps (accuracy deltas up to 16%, Script Consistency Rates as low as 24.8%) and limited cross-script transferability. (3) **Explainability through Divergence**: Evidence that script choice functions as a latent trigger, causing models to adopt divergent reasoning strategies despite constant visual evidence.

## 2 Related Work

The ascent of Vision-Language Models (VLMs), from CLIP's foundational image-text alignment (Radford et al., 2021) to instruction-tuned systems like LLaVA (Liu et al., 2024) and Qwen-VL (Bai et al., 2023), has been accompanied by claims of multilingual competence across dozens of languages. Yet beneath this veneer lies a critical blind spot: the **"One Language, One Script" (OLOS) paradigm**. Current benchmarks conflate language with orthography, treating script as deterministic rather than variable. For over one billion speakers of digraphic languages like Punjabi (ਗੁਰਮੁਖੀ (Gurmukhi)), شاہمکھی (Shahmukhi), Roman), Serbian (Cyrillic, Latin), and Kurdish (Arabic, Latin, Cyrillic), this assumption masks orthographic bias that fragments AI access.

### 2.1 Multilingual Multimodal Benchmarks: Breadth Without Orthographic Depth

Recent benchmarks have achieved unprecedented linguistic scale. IGLUE (Bugliarello et al., 2022) pioneered cross-lingual transfer across 20 languages, while MaRVL (Liu et al., 2021) exposed Western-centric biases through native-speaker concepts in five languages. XM3600 (Thapliyal et al., 2022) scaled captioning to 36 languages, PaLo (Maaz et al., 2024) evaluated instruction-following across 50+ languages, and MVL-SIB (Schmidt et al., 2025) assessed 205 languages for topical matching. Most recently, BLEnD-Vis (Tan et al., 2025) advanced cultural evaluation across 16 regions, while IndicVisionBench (Faraz et al., 2025) and ALM-Bench (Vayani et al., 2025) expanded to 10 Indian and 100 global languages respectively. However, these milestones share a fundamental limitation: *each language appears in exactly one script*. MaRVL represents Tamil in Tamil script and Swahili in Latin, but never tests whether Tamil speakers using Romanization experience equivalent performance. This OLOS assumption artificially inflates multilingual capabilities, a model achieving 85% on "Punjabi" in Gurmukhi may drop to 69% in Shahmukhi for identical content, a disparity invisible on current leaderboards.

### 2.2 The Script Gap: From Text-Only Evidence to Multimodal Urgency

Text-only NLP provides clear evidence of script-dependent degradation. Pfeiffer et al. (2021) showed multilingual BERT fails catastrophically on unseen scripts, while Rust et al. (2021) proved tokenizers allocate subword capacity inequitably, with low-resource scripts receiving 3–5× fewer tokens per unit. This "Script Gap" has real-world consequences: Khullar et al. (2025) found Romanized Hindi-Urdu queries in Indian healthcare triage degraded F1 scores by 5–12 points, with token-level entropy spiking from 1.7 to 4.6 bits, potentially affecting 2 million maternal health cases annually. While Amrhein and Sennrich (2020) showed Romanization strips phonological nuance and Nguyen et al. (2024) demonstrated explicit cross-script training can mitigate this for CJKV languages, *no prior work extends script-gap analysis to multimodal reasoning*, leaving unanswered whether visual grounding compensates for orthographic fragility.

### 2.3 Cultural Grounding and Script-Locked Knowledge

Effective multimodal reasoning demands cultural grounding beyond object recognition (Yin et al., 2021). MaRVL (Liu et al., 2021) exposed failures on non-Western concepts, while BLEnD-Vis (Tan et al., 2025) revealed only 42.19% joint correctness between textual and visual cultural knowledge. Yet no benchmark probes whether cultural knowledge is *script-locked*. This omission is critical: in Punjabi, Gurmukhi evokes Sikh contexts, Shahmukhi aligns with Islamic spheres, and Roman signals diaspora hybridity. PuMVR's culturally-grounded categories enable testing whether models activate distinct knowledge pathways based on script alone.

### 2.4 OCR, CoT, and Cross-Modal Script Integration

Scene text reasoning remains challenging, with accuracy dropping 15–30% for non-Latin scripts (Singh et al., 2019; Biten et al., 2019; Rahmanzadehgervi et al., 2025). However, prior work evaluates OCR in isolation, no benchmark systematically varies image text script independently from query script. PuMVR's 3×3 factorial design enables measuring the Script Disparity Index (SDI): the performance penalty when orthographies misalign across modalities. Chain-of-Thought (CoT) analysis (Wei et al., 2023) has emerged as a window into model reasoning. In multimodal settings, Wang et al. (2024) showed CoT reveals reasoning patterns, but we are the first to demonstrate script functions as a latent trigger altering CoT pathways. When queried in Gurmukhi about a festival image, gpt-4o focuses on Sikh cultural markers; the identical Shahmukhi query shifts to Islamic features, a form of script-modulated visual confirmation bias (Vo et al., 2025).

### 2.5 Positioning PuMVR

Unlike concurrent efforts prioritizing language breadth, PuMVR prioritizes *orthographic depth*. We isolate script as an independent variable across 375 parallel instances and six reasoning categories, enabling novel metrics: Script Consistency Rate (SCR) measures cross-script correctness, Transfer Efficiency (TE) quantifies few-shot transferability, and Script Disparity Index (SDI) captures OCR misalignment penalties. Our findings, accuracy deltas reaching 16%, SCR as low as 27%, and systematic reasoning divergence, establish that current "multilingual" VLMs are not truly multi-script, challenging the OLOS paradigm and providing a methodological blueprint for orthographic equity in global AI.

## 3 The PuMVR Benchmark

We introduce **PuMVR** (Punjabi Multimodal Visual Reasoning), the first benchmark designed to isolate script as an independent variable in multimodal reasoning. Unlike traditional multilingual benchmarks that conflate language with orthography, PuMVR provides 375 culturally grounded image-reasoning tasks, each existing in perfect semantic equivalence across ਗੁਰਮੁਖੀ (Gurmukhi), شاہمکھی (Shahmukhi), and Roman scripts.

### 3.1 Design Rationale: Why Punjabi?

Punjabi serves as an ideal diagnostic for script bias due to three critical properties.

- Its three active scripts, Gurmukhi (Indian Punjab, 50M speakers), Shahmukhi (Pakistani Punjab, 60M speakers), and Roman (diaspora/digital, 15M users), are geographically and culturally isolated, minimizing training data overlap.
- The scripts represent fundamentally different visual systems: Gurmukhi is an Indic abugida, Shahmukhi uses Perso-Arabic cursive, and Roman relies on Latin linearity.
- Punjabi remains low-resource compared to Hindi or Urdu, amplifying reliance on memorized patterns over semantic grounding.

Crucially, certain cultural concepts carry script-specific associations, Gurmukhi predominates in Sikh contexts (Golden Temple), while Shahmukhi aligns with Islamic heritage (Badshahi Mosque), enabling tests of script-locked knowledge retrieval.

### 3.2 Dataset Composition and Structure

PuMVR comprises 375 instances distributed across six reasoning categories (App. B) de-signed to test distinct multimodal capabilities, as shown in Table 1. Each instance contains: an image, question text in all three scripts, four multiple-choice options per script, the correct answer per script, and a human-authored reasoning explanation.

| Category | Count | Description |
|---|---|---|
| Visual Analogies | 60 | Relational reasoning |
| Cultural Object Recognition | 60 | Punjabi artifact recognition |
| Festival & Celebration Reasoning | 60 | Cultural knowledge and temporal reasoning |
| Architectural & Landmark Recognition | 60 | Visual and geographic grounding |
| Text-in-Image Reasoning | 70 | Cross-Script OCR Reasoning |
| Abstract Visual-Linguistic Reasoning | 65 | Basic spatial and logical reasoning with labels |
| **Total** | **375** | |

Table 1: Distribution of PuMVR instances across categories.

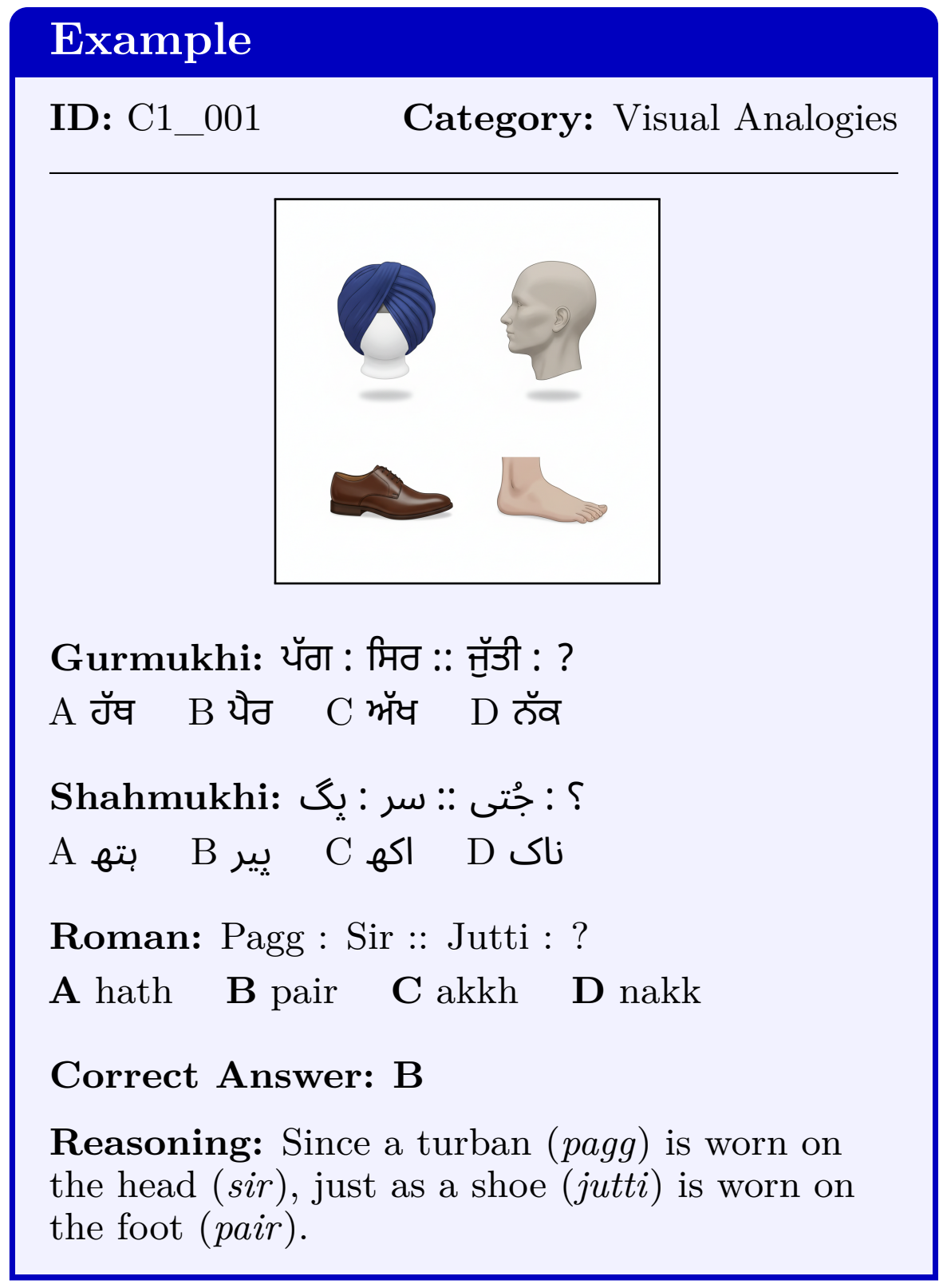

Figure 1: Sample instance from PuMVR showing cross-script equivalent options.

### 3.3 Quality Assurance

All 375 instances were created by a team of two native Punjabi speakers: one researcher fluent in Gurmukhi and Roman scripts, and one consultant fluent in Shahmukhi and Gurmukhi. Initial instances were authored in Gurmukhi and Roman, with Shahmukhi translations and cultural verification provided by the consultant. This process ensured semantic equivalence, cultural authenticity, and correctness across all three orthographic systems.

### 3.4 From Language Coverage to Orthographic Equity

PuMVR represents a paradigm shift from language *breadth* to orthographic *depth*. A model achieving 90% on **Punjabi** exclusively in Gurmukhi is not multilingual, it is uniscriptual. By making script an explicit, controllable variable, PuMVR enables diagnosing bias origins, quantifying the **Script Gap,** and developing equitable training strategies. Our methodology extends to dozens of multiscript languages (Hindi-Urdu, Serbian, Kurdish, Sindhi), providing a blueprint for orthographic equity in global AI.

## 4 Experimental Setup

Our experimental framework systematically interrogates the **Script-Reality Gap** through five complementary investigations, each designed to expose a distinct facet of orthographic bias in Vision-Language Models.

We evaluate 10 state-of-the-art VLMs spanning proprietary frontier systems and open-weights alternatives (Table 2), enabling us to audit industry-leading systems while providing reproducible baselines.

### 4.1 Experiment 1: The "Script Gap" Quantification

**Objective**: Establish the existence and magnitude of script-dependent performance bias under identical semantic conditions.

**Design**: Each PuMVR instance is evaluated in three isolated passes, one per script, to prevent cross-script priming. Models receive the image, question, and four options in a single script using script-specific instruction templates (App. D), and must output the exact text of the correct option. We evaluate model performance using the following metrics.

| Model | Organization |
|---|---|
| **Proprietary Frontier Models** | |
| gpt-4o | OpenAI |
| gemini-2.5-flash | Google DeepMind |
| claude-sonnet-4 | Anthropic |
| grok-4-1-fast-reasoning | xAI |
| **Open-Weights Models** | |
| Qwen2-VL-7B-Instruct | Alibaba |
| Qwen2-VL-72B-Instruct | Alibaba |
| Llama-3.2-11B-Vision | Meta |
| LLaVA-OneVision-1.5-8B-Instruct | LMMS-Lab |
| InternVL2_5-26B | OpenGVLab |
| Kimi-VL-A3B-Instruct | Moonshot AI |

Table 2: Evaluated VLMs spanning frontier and open-weights systems.

1. **Script Accuracy** ($\text{Acc}_s$): Raw per-script performance

$$\text{Acc}_s = \frac{1}{|I|} \sum_{i \in I} \mathbb{1}[\text{Pred}_s(i) = \text{GT}_s(i)] \quad (1)$$

2. **Script Consistency Rate (SCR)**: Percentage of instances answered correctly across all three scripts simultaneously

$$\text{SCR} = \frac{1}{|I|} \sum_{i \in I} \prod_{s \in \mathcal{S}} \mathbb{1}[\text{Correct}_s(i)] \quad (2)$$

where $\mathcal{S} = \{\text{Gur}, \text{Shah}, \text{Rom}\}$
SCR serves as a strict script-agnostic benchmark. A model achieving 90% per-script accuracy but only 70% SCR reveals that 20% of its knowledge is orthographically fragmented.

3. **Performance Delta** ($\Delta$): Maximum accuracy variance

$$\Delta = \max_{s_1, s_2 \in S} |\text{Acc}_{s_1} - \text{Acc}_{s_2}| \quad (3)$$

## 4.2 Experiment 2: Modality Importance Ablation

**Objective**: Determine whether visual grounding compensates for weak script comprehension or merely provides additive benefit.

**Design**: We compare Text-Only (question + options, no image) versus Multimodal (complete input) conditions across all scripts.

**Metric**: Visual Gain (VG) quantifies visual contribution:

$$\text{VG}_s = \text{Acc}_{\text{Multimodal}}(s) - \text{Acc}_{\text{Text-Only}}(s) \quad (4)$$

Uniform VG across scripts indicates visual information does not close the script gap, the bias is systematic, not compensatory.

## 4.3 Experiment 3: Cross-Script Transfer with Few-Shot Learning

**Objective**: Test whether reasoning patterns transfer across orthographies or remain script-locked.

**Design**: We employ $k = 3$ in-context exemplars under three conditions:

1. **Monoscript**: Examples and test in same script (e.g., Gurmukhi → Gurmukhi)
2. **Cross-Script**: Examples in different script (e.g., Roman → Gurmukhi)
3. **Mixed-Script**: Examples rotated through all three scripts

**Metrics**:

1. **Few-Shot Lift (FSL)**: Improvement from zero-shot

$$\text{FSL}_s = \text{Acc}_{\text{fs}}(s \to s) - \text{Acc}_{\text{zs}}(s) \quad (5)$$

where $\text{Acc}_{\text{fs}}$ and $\text{Acc}_{\text{zs}}$ denote the accuracies for few-shot and zero-shot conditions, respectively.

2. **Transfer Efficiency (TE)**: Cross-script to in-script ratio

$$\text{TE}_{T \to S} = \frac{\text{Acc}_{\text{Few-Shot}}(T \to S)}{\text{Acc}_{\text{Few-Shot}}(S \to S)} \times 100\% \quad (6)$$

TE quantifies how much reasoning intelligence survives script translation. If TE $<$ 50%, the model has not learned the concept, it has merely memorized visual-textual patterns specific to one script. If TE exhibits asymmetry (e.g., $\text{TE}_{\text{Roman} \to \text{Gurmukhi}} \gg \text{TE}_{\text{Gurmukhi} \to \text{Roman}}$), it reveals an anchor script bias.

## 4.4 Experiment 4: OCR and Script Mismatch Robustness

**Objective**: Quantify the performance penalty when image-text and question scripts are misaligned.

**Design**: Using Category 5, we create a 3×3 factorial design varying image text script (Gurmukhi/Roman/Shahmukhi) and question script independently, yielding 9 conditions.

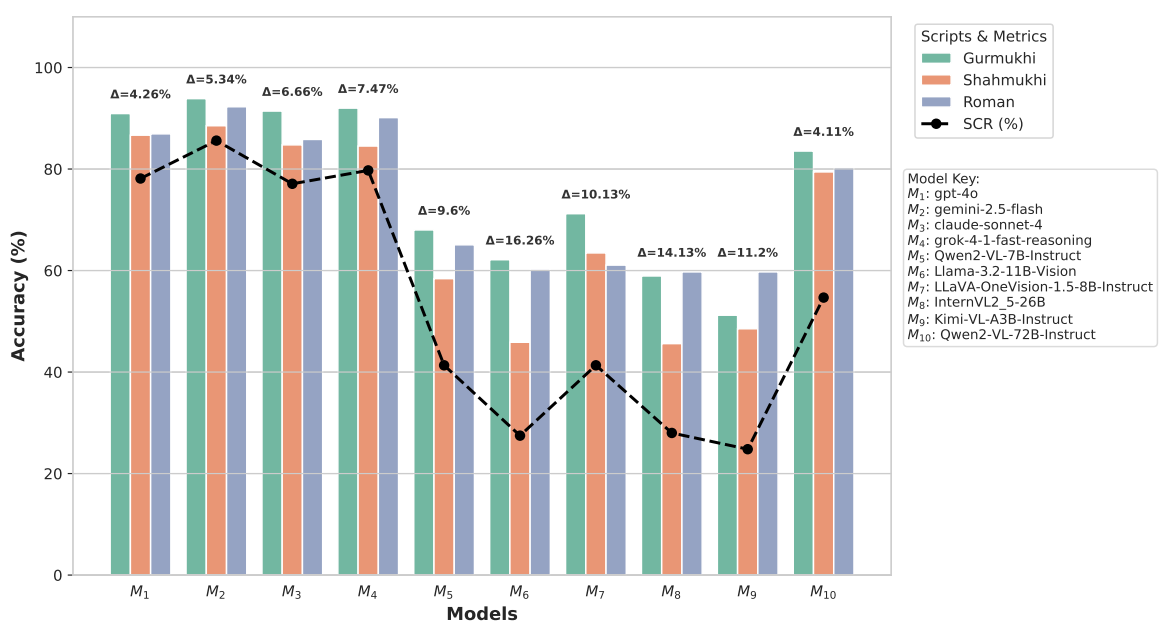

Figure 2: Script-dependent accuracy and Script Consistency Rate (SCR) across 10 VLMs.

**Metric**: Script Disparity Index (SDI) measures maximum performance spread:

$$\text{SDI}_I = \max_{Q \in S} \text{Acc}(I, Q) - \min_{Q \in S} \text{Acc}(I, Q) \quad (7)$$

High SDI indicates script-brittle reasoning, models cannot abstract away from orthographic surface forms.

### 4.5 Experiment 5: Script-Induced Reasoning Divergence

**Objective**: Investigate whether script variation correlates with different reasoning verbalizations through Chain-of-Thought analysis.

**Design**: We employ structured CoT prompting on 225 stratified instances (60% of PuMVR). Each instance is evaluated with CoT prompts in all three scripts. The prompt enforces a 5-step structure: (1) Visual Grounding, (2) Linguistic Parsing, (3) Relation Abstraction, (4) Cross-Modal Integration, (5) Deduction.

**Analysis**: We measure accuracy with CoT and compute the delta from Experiment 1's zero-shot baseline:

$$\Delta_{\text{CoT}}(s) = \text{Acc}_{\text{CoT}}(s) - \text{Acc}_{\text{Zero-Shot}}(s) \quad (8)$$

Negative deltas indicate that structured reasoning prompts destabilize script-specific pathways, while positive deltas suggest stabilization of existing strategies. Systematic variance in $\Delta_{\text{CoT}}$ across scripts would reveal orthographically fragmented logic rather than unified semantic reasoning.

## 5 Results and Analysis

Our systematic evaluation of 10 state-of-the-art VLMs across PuMVR's 375 instances exposes a fundamental systematic limitation: **current multilingual systems are not truly multi-script**. We present evidence that orthographic variation functions as a latent trigger, systematically fragmenting model reasoning despite constant semantic content and visual grounding.

| Model | Gurmukhi | | Shahmukhi | | Roman | |
|---|---|---|---|---|---|---|
| | T | VG | T | VG | T | VG |
| gpt-4o | 74.1 | 16.8 | 69.9 | 16.8 | 62.4 | 24.5 |
| gemini-2.5-flash | 71.2 | 22.7 | 67.7 | 20.8 | 66.9 | 25.3 |
| claude-sonnet-4 | 74.3 | 17.2 | 64.5 | 20.2 | 58.4 | 27.4 |
| grok-4-1-fast-reasoning | 73.6 | 18.4 | 63.2 | 21.3 | 67.2 | 22.9 |
| Qwen2-VL-7B-Instruct | 40.8 | 27.2 | 37.1 | 21.3 | 34.9 | 30.1 |
| Llama-3.2-11B-Vision | 46.9 | 15.2 | 38.4 | 7.5 | 34.7 | 25.6 |
| LLaVA-OneVision-1.5-8B-Instruct | 48.8 | 22.4 | 45.3 | 18.1 | 37.1 | 24.0 |
| InternVL2_5-26B | 36.8 | 22.1 | 32.3 | 13.3 | 32.8 | 26.9 |
| Kimi-VL-A3B-Instruct | 38.1 | 13.1 | 37.1 | 11.5 | 35.7 | 24.0 |
| Qwen2-VL-72B-Instruct | 47.5 | 36.1 | 53.1 | 26.4 | 51.7 | 28.5 |

Table 3: Modality ablation showing Text-Only accuracy (T) and Visual Gain (VG). Blue indicates highest values, red indicates lowest values per column.

### 5.1 The Script Gap is Universal and Substantial

Figure 2 reveals our primary finding: **all evaluated models exhibit script-dependent performance degradation**, detailed in the per-model breakdown (App. C). Accuracy deltas range from 4.11% (Qwen2-VL-72B-Instruct) to 16.26% (Llama-3.2-11B-Vision). Critically, this affects frontier models, gpt-4o, gemini-2.5-flash, and claude-sonnet-4 demonstrate $\Delta$ values of 4.26%, 5.34%, and 6.66% respectively, proving script bias is *not* a low-resource artifact but a systematic limitation of current VLM designs.

The **Script Consistency Rate (SCR)** exposes the severity of orthographic fragmentation. gpt-4o achieves 90.93% in Gurmukhi yet records only 78.13% SCR, **12.8% of its knowledge is orthographically locked**. This pattern intensifies in open-weights models: Llama-3.2-11B-Vision's SCR of 27.47% means nearly three-quarters of instances cannot be solved consistently across scripts, rendering reported Punjabi capability meaningless for real-world multi-script users.

**Statistical Significance of Script Gaps.** To verify that observed performance gaps exceed chance variation, we apply McNemar's test to all pairwise script comparisons for each model, treating per-instance correctness as paired binary observations across 375 instances (Table 4). Even the smallest deltas in frontier models are statistically significant: GPT-4o's 4.0% Gurmukhi–Shahmukhi gap ($p$=0.025) and Claude Sonnet 4's 5.87% gap ($p$<0.001) confirm these differences are not attributable to sampling noise. Grok and Claude show highly significant Gurmukhi–Shahmukhi disparities ($p$<0.001), while Shahmukhi consistently underperforms across models. The sole exception is Qwen2-VL-72B-Instruct, whose gaps do not reach significance ($p$>0.13 across all pairs), suggesting more balanced cross-script representation in its training data. These results confirm the Script Gap is a statistically robust phenomenon, not an artifact of dataset size.

| Model | Script Pair | $p$-value | Sig. |
|---|---|---|---|
| gpt-4o | Gur vs Shah | 0.025 | * |
| gemini-2.5-flash | Gur vs Shah | 0.001 | ** |
| claude-sonnet-4 | Gur vs Shah | <0.001 | *** |
| grok-4-1-fast-reasoning | Gur vs Shah | <0.001 | *** |
| Llama-3.2-11B-Vision | Gur vs Shah | <0.001 | *** |
| Qwen2-VL-72B-Instruct | All pairs | >0.13 | ns |

Table 4: McNemar's test (Gurmukhi vs. Shahmukhi, the largest gap pair). Full results in App. E.

Eight of ten models peak in Gurmukhi, reflecting its predominance in Indian web corpora. Shahmukhi consistently underperforms, averaging 7.7% below Gurmukhi, indicating Perso-Arabic cursive representations remain systematically undertrained even in frontier systems.

### 5.2 Visual Grounding Provides Parallel, Not Compensatory Gain

Table 3 reveals that visual information boosts performance uniformly (VG: 7.5–36.1%) but **does not close the script gap**. gpt-4o's VG is nearly identical across Gurmukhi (16.8%) and Shahmukhi (16.8%), yet absolute accuracies differ by 4.3%. Images provide **additive benefit**, not **compensatory repair**, the underlying orthographic fragmentation persists in multimodal settings.

Roman script exhibits highest VG (avg: 25.9%) despite often being the best-performing script in zero-shot settings, suggesting models rely on memorized patterns in high-resource scripts but require visual evidence when orthographic priors weaken. Llama-3.2-11B-Vision's substantially low Shahmukhi VG (7.5%) indicates **script-specific visual grounding failure**: even with images, it cannot integrate Perso-Arabic text effectively.

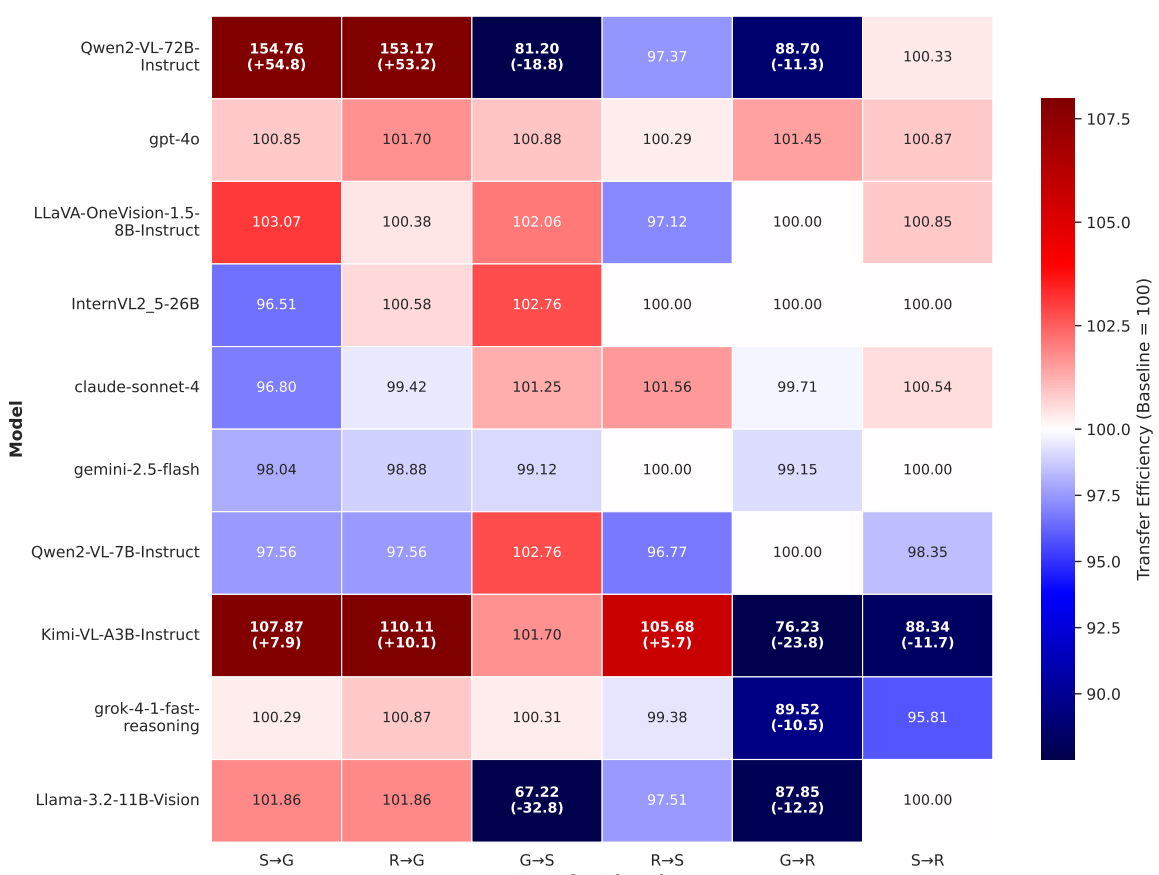

Figure 3: Transfer Efficiency (TE) heatmap (baseline=100%). Values above 100 indicate cross-script examples outperform monoscript; below 100 indicates reasoning loss.

### 5.3 Limited Cross-Script Transfer Suggests Script-Dependent Knowledge

Table 6 and Figure 3 reveal that reasoning patterns do not transfer across scripts, indicating knowledge may be orthographically fragmented.

**Negative Few-Shot Lift Exposes Brittleness.** Four models exhibit *performance degradation* when provided examples. Qwen2-VL-72B-Instruct's G→G FSL of −**49.9%** is catastrophic: zero-shot accuracy of 83.5% collapses to 33.6% with three Gurmukhi examples, suggesting fragile reasoning where in-context learning triggers interference.

**Transfer Efficiency Reveals Anchor-Script Asymmetry.** Frontier models achieve near-perfect TE (98–101%), indicating script-agnostic grounding. However, Qwen2-VL-72B-Instruct exhibits extreme asymmetry: $TE_{S \to G}$=154.76%, $TE_{R \to G}$=153.17% (cross-script outperforms in-script), yet $TE_{G \to S}$=81.20% (19% reasoning loss). Llama-3.2-11B-Vision's $TE_{G \to S}$=67.22% represents **33% intelligence loss**, strong evidence that knowledge exhibits script-specific dependen-cies.

### 5.4 OCR Script Mismatch Amplifies Fragility

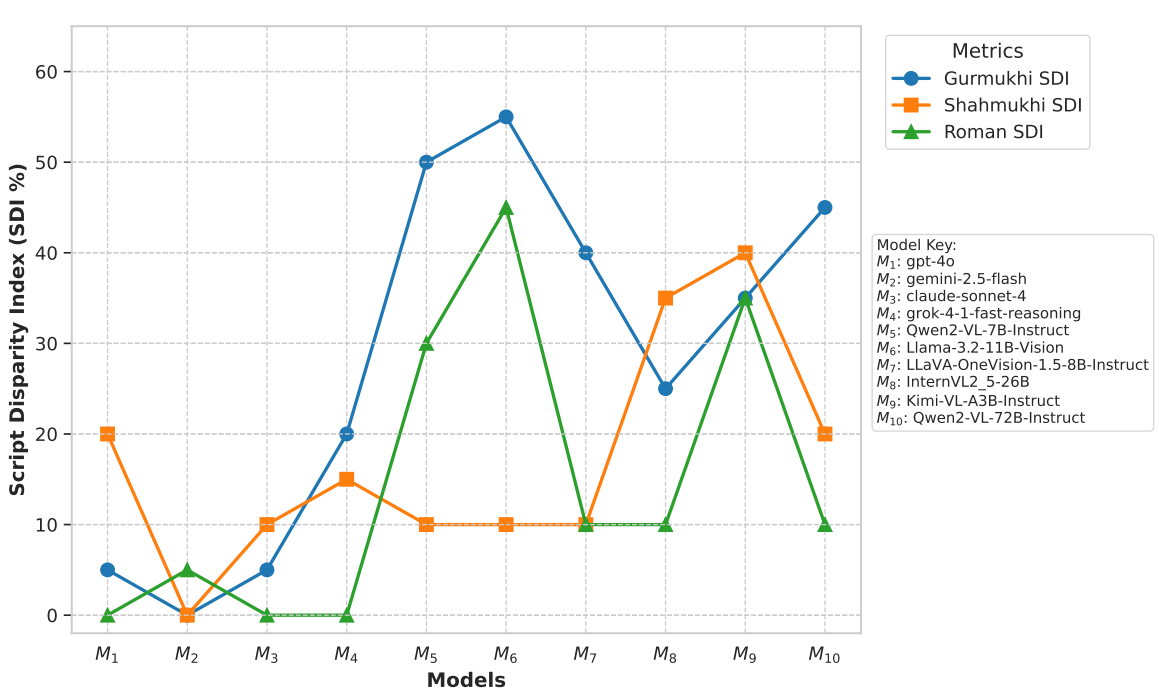

Figure 4: Script Disparity Index (SDI).

Figure 4 quantifies Script Disparity Index (SDI), performance penalties when image-text and question scripts misalign. Frontier models exhibit low SDI (0–20%), with three out of four models achieving SDI=0% for Roman images. In contrast, Llama-3.2-11B-Vision records $\text{SDI}_{\text{Gurmukhi}}$=55% and Qwen2-VL-7B-Instruct shows $\text{SDI}_{\text{Gurmukhi}}$=50%, revealing deployment considerations for multi-script documents. All models except gemini-2.5-flash exhibit elevated Shahmukhi SDI (15–40%), indicating systematic Perso-Arabic OCR failures.

### 5.5 Script Correlates with Reasoning Strategy Variation

| Model | Gur. | Shah. | Rom. | Δ |
|---|---|---|---|---|
| gpt-4o | $91.5_{+0.6}$ | $66.2_{-20.4}$ | $80.0_{-6.9}$ | 25.3 |
| gemini-2.5-flash | $59.5_{-34.4}$ | $88.9_{+0.4}$ | $84.5_{-7.7}$ | 29.4 |
| claude-sonnet-4 | $88.8_{-2.6}$ | $73.8_{-11.0}$ | $69.6_{-16.2}$ | 19.2 |
| grok-4-1-fast-reasoning | $63.1_{-28.9}$ | $79.9_{-4.6}$ | $83.0_{-7.2}$ | 19.9 |
| Qwen2-VL-7B-Instruct | $57.5_{-10.5}$ | $50.9_{-7.5}$ | $44.0_{-21.1}$ | 13.5 |
| Llama-3.2-11B-Vision | $51.9_{-10.3}$ | $40.9_{-5.0}$ | $43.6_{-16.7}$ | 11.0 |
| LLaVA-OneVision-1.5-8B-Instruct | $69.4_{-1.8}$ | $52.5_{-11.0}$ | $45.8_{-15.3}$ | 23.6 |
| InternVL2_5-26B | $50.5_{-8.4}$ | $41.2_{-4.4}$ | $49.1_{-10.6}$ | 9.3 |
| Kimi-VL-A3B-Instruct | $49.8_{-1.4}$ | $42.9_{-5.7}$ | $40.4_{-19.3}$ | 9.4 |
| Qwen2-VL-72B-Instruct | $86.7_{+3.1}$ | $76.9_{-2.5}$ | $76.2_{-4.0}$ | 10.5 |

Table 5: CoT-induced reasoning divergence (subscripts: delta from zero-shot). Blue indicates highest and red indicates lowest values per column.

Table 5 reveals that structured CoT prompting exhibits **script-selective destabilization**. gpt-4o's Shahmukhi accuracy drops 20.4% under CoT while Gurmukhi remains stable (+0.6%), suggesting Shahmukhi reasoning relies on implicit heuristics that collapse under explicit articulation. Most dramatically, **gemini-2.5-flash's Gurmukhi accuracy plummets 34.4%** (93.9%→59.5%) while Shahmukhi stabilizes, revealing inverted script preferences under cognitive load.

Each model exhibits a stable script where CoT reinforces reasoning (Gurmukhi for gpt-4o/claude-sonnet-4; Shahmukhi for gemini-2.5-flash), likely reflecting dominant orthography in pre-training. Six models show uniformly negative deltas (avg: 10%), potentially reflecting task difficulty or sensitivity to CoT prompting. While consistent with shallow pattern-matching, these results may also reflect different verbalization strategies across scripts or training data imbalances.

## 6 Conclusion

We expose a critical blind spot in multilingual VLM evaluation: the "One Language, One Script" paradigm. Through PuMVR's 375 parallel-script instances, we provide the first systematic evidence that state-of-the-art models exhibit substantial script-dependent bias - accuracy deltas reach 16%, Script Consistency Rates fall to 24.8%, and Transfer Efficiency drops below 67%. Critically, visual grounding provides additive benefit but does not close the script gap, proving the bias is systematic. Chain-of-Thought analysis reveals that script functions as a latent cognitive trigger, causing models to adopt divergent reasoning pathways despite constant visual evidence.

These findings suggest current multilingual claims may not generalize uniformly across writing systems. We propose Script Consistency Rate as essential for script-agnostic evaluation and present a replicable methodology applicable to dozens of multi-script languages. Future work should investigate orthographic equity to serve the billion-plus users of multi-script languages with equal fidelity. Achieving equitable AI requires moving beyond language counts toward orthographic depth.

| Model | Monoscript | | | Cross-Script | | | Mixed-Script | | |
|---|---|---|---|---|---|---|---|---|---|
| | G→G | S→S | R→R | →G | →S | →R | →G | →S | →R |
| gpt-4o | 93.9 +2.9 | 90.7 +4.0 | 92.0 +5.1 | 95.1 +4.1 | 91.2 +4.5 | 93.1 +6.1 | 94.7 +3.7 | 90.4 +3.7 | 92.5 +5.6 |
| gemini-2.5-flash | 95.5 +1.6 | 90.4 +1.9 | 93.6 +1.3 | 94.0 +0.1 | 90.0 +1.5 | 93.2 +0.9 | 94.4 +0.5 | 89.3 +0.8 | 94.7 +2.4 |
| claude-sonnet-4 | 91.7 +0.3 | 85.6 +0.8 | 90.7 +4.8 | 90.0 −1.4 | 86.8 +2.0 | 90.8 +4.9 | 90.4 −1.0 | 86.7 +1.9 | 90.1 +4.3 |
| grok-4-1-fast-reasoning | 91.5 −0.5 | 85.9 +1.3 | 89.1 −1.1 | 92.0 0.0 | 85.7 +1.2 | 82.5 −7.6 | 93.1 +1.1 | 85.3 +0.8 | 82.4 −7.7 |
| Qwen2-VL-7B-Instruct | 65.6 −2.4 | 57.9 −0.5 | 64.5 −0.5 | 64.0 −4.0 | 57.7 −0.7 | 64.0 −1.1 | 65.6 −2.4 | 57.6 −0.8 | 62.4 −2.7 |
| Llama-3.2-11B-Vision | 71.7 +9.6 | 64.3 +18.4 | 65.9 +5.6 | 73.1 +10.9 | 52.9 +7.1 | 61.9 +1.6 | 70.7 +8.5 | 56.5 +10.7 | 65.1 +4.8 |
| LLaVA-OneVision-1.5-8B-Instruct | 69.6 −1.6 | 64.8 +1.3 | 62.7 +1.6 | 70.8 −0.4 | 64.5 +1.1 | 62.9 +1.9 | 69.9 −1.3 | 65.1 +1.6 | 63.5 +2.4 |
| InternVL2_5-26B | 45.9 −13.1 | 38.7 −6.9 | 53.3 −6.4 | 45.2 −13.7 | 39.2 −6.4 | 53.3 −6.4 | 45.6 −13.3 | 37.1 −8.5 | 52.3 −7.5 |
| Kimi-VL-A3B-Instruct | 47.5 −3.7 | 46.9 −1.6 | 59.5 −0.3 | 51.7 +0.5 | 48.7 +0.1 | 48.9 −10.8 | 49.9 −1.3 | 49.1 +0.5 | 47.2 −12.5 |
| Qwen2-VL-72B-Instruct | 33.6 −49.9 | 70.9 −8.5 | 80.3 0.0 | 51.7 −31.8 | 63.3 −16.1 | 75.9 −4.4 | 47.5 −36.1 | 70.1 −9.3 | 78.7 −1.6 |

Table 6: Few-Shot Lift (FSL): accuracy with $k = 3$ examples (subscripts show lift from zero-shot in gray). Blue indicates highest accuracy, red indicates lowest accuracy per column.

## Limitations

**Dataset Scope.** Our benchmark contains 375 curated instances across six categories. While this represents the first systematic cross-script multimodal evaluation, the size constrains fine-grained statistical analysis and represents a focused subset of reasoning capabilities. We prioritized quality and semantic equivalence over scale; future work should expand instance counts while maintaining our rigorous parallel-script methodology.

**Linguistic Generalizability.** We focus on Punjabi, whose three scripts provide typological diversity and minimal corpus overlap. However, script bias patterns may differ for: (1) languages with closer orthographic relationships (Serbian Cyrillic/Latin), (2) logographic systems (Chinese traditional/simplified), or (3) scripts with more balanced web representation. Our methodology is transferable, but empirical validation across diverse language families is needed before broad generalization. We provide a blueprint, not a universal law.

**Evaluation Settings.** We assess zero-shot and few-shot performance without fine-tuning. Script bias may behave differently under full fine-tuning with script-balanced data, instruction tuning for cross-script robustness, or retrieval-augmented architectures. Our findings reflect contemporary deployment settings but may not predict behavior under targeted mitigation strategies.

**Chain-of-Thought as Proxy.** Our CoT analysis treats verbalized reasoning as a window into model behavior, acknowledging it may reflect post-hoc rationalization rather than actual decision processes. Observed divergence could stem from language-specific discourse conventions rather than fundamental reasoning differences. We interpret CoT outputs as evidence of observable behavioral patterns, scripts correlate with different reasoning verbalizations, while remaining cautious about claims regarding internal computation without direct representation probing.

**Model Coverage.** We evaluate 10 contemporary VLMs representing current state-of-the-art. Findings may not generalize to future architectures with improved cross-script mechanisms, proprietary systems with undisclosed script-balancing, or domain-specific models. Our work establishes a baseline for measuring progress rather than immutable limitations.

**Cultural Entanglement.** Despite rigorous efforts at semantic equivalence, some instances may carry subtle script-specific cultural associations (Gurmukhi/Sikh, Shahmukhi/Islamic contexts) that are difficult to fully disentangle in a culturally grounded benchmark. This reflects the lived reality of script-culture associations but may introduce confounds beyond pure orthographic variation.

## Ethics Statement

**Cultural Authenticity and Representation.** PuMVR was developed by native Punjabi speakers fluent in all three scripts and reviewed by community members. This ensured linguistic accuracy, cultural authenticity, and avoidance of stereotyping. Our teams lived experience with script-dependent technological barriers informed the work but may limit per-spective on other multi-script contexts.

**Data Sources and Privacy.** All images are either AI-generated, public domain, or openly licensed. No personally identifiable information is included. Human subjects in photographs are from public historical archives or are AI-generated.

**Dual-Use Considerations.** This research exposes systemic bias to promote equitable AI access for multi-script communities. However, we acknowledge potential misuse: script-dependent performance insights could facilitate targeted disinformation or discrimination against low-resource script users. Our work diagnoses rather than optimizes such biases; we advocate for script-agnostic improvements that benefit all orthographies equally. We believe the benefits, improved fairness for over a billion users, substantially outweigh the risks.

**Environmental Impact.** Total evaluation required approximately 250 Nvidia GH200 GPU hours, consuming an estimated 193 kWh of electricity and producing about 91.4 kg $CO_2$ (equivalent to 368 km of passenger vehicle travel).[1] While non-trivial, this cost is reported transparently and reflects current practices for large-scale multimodal evaluation, with potential benefits for improving equity across over one billion users of multi-script languages.

**Transparency.** We release all code and detailed experimental protocols at https://github.com/prabhjotschugh/PuMVR-benchmark-for-VLMs to enable reproducibility and community validation of our findings.

[1]Estimates assume a 700W average power draw for the GH200 Superchip and a PUE of 1.1.

## References

Chantal Amrhein and Rico Sennrich. 2020. On Romanization for model transfer between scripts in neural machine translation. In *Findings of the Association for Computational Linguistics: EMNLP 2020*, pages 2461–2469, Online. Association for Computational Linguistics.

Jinze Bai, Shuai Bai, Shusheng Yang, Shijie Wang, Sinan Tan, Peng Wang, Junyang Lin, Chang Zhou, and Jingren Zhou. 2023. Qwen-vl: A versatile vision-language model for understanding, localization, text reading, and beyond. *Preprint*, arXiv:2308.12966.

Ali Furkan Biten, Ruben Tito, Andres Mafla, Lluis Gomez, Marçal Rusiñol, Ernest Valveny, C. V. Jawahar, and Dimosthenis Karatzas. 2019. Scene text visual question answering. *Preprint*, arXiv:1905.13648.

Emanuele Bugliarello, Fangyu Liu, Jonas Pfeiffer, Siva Reddy, Desmond Elliott, Edoardo Maria Ponti, and Ivan Vuli. 2022. Iglue: A benchmark for transfer learning across modalities, tasks, and languages. *Preprint*, arXiv:2201.11732.

Soravit Changpinyo, Linting Xue, Michal Yarom, Ashish V. Thapliyal, Idan Szpektor, Julien Amelot, Xi Chen, and Radu Soricut. 2023. Maxm: Towards multilingual visual question answering. *Preprint*, arXiv:2209.05401.

Ali Faraz, Akash, Shaharukh Khan, Raja Kolla, Akshat Patidar, Suranjan Goswami, Abhinav Ravi, Chandra Khatri, and Shubham Agarwal. 2025. Indicvisionbench: Benchmarking cultural and multilingual understanding in vlms. *Preprint*, arXiv:2511.04727.

Manurag Khullar, Utkarsh Desai, Poorva Malviya, Aman Dalmia, and Zheyuan Ryan Shi. 2025. Script gap: Evaluating llm triage on indian languages in native vs roman scripts in a real world setting. *Preprint*, arXiv:2512.10780.

Fangyu Liu, Emanuele Bugliarello, Edoardo Maria Ponti, Siva Reddy, Nigel Collier, and Desmond Elliott. 2021. Visually grounded reasoning across languages and cultures. In *Proceedings of the 2021 Conference on Empirical Methods in Natural Language Processing*, pages 10467–10485, Online and Punta Cana, Dominican Republic. Association for Computational Linguistics.

Haotian Liu, Chunyuan Li, Yuheng Li, Bo Li, Yuanhan Zhang, Sheng Shen, and Yong Jae Lee. 2024. LLaVA-NeXT: Improved reasoning, OCR, and world knowledge. `https://llava-vl.github.io/blog/2024-01-30-llava-next/`.

Muhammad Maaz, Hanoona Rasheed, Abdelrahman Shaker, Salman Khan, Hisham Cholakal, Rao M. Anwer, Tim Baldwin, Michael Felsberg, and Fahad S. Khan. 2024. Palo: A polyglot large multimodal model for 5b people. *Preprint*, arXiv:2402.14818.

Hoang Nguyen, Chenwei Zhang, Ye Liu, Natalie Parde, Eugene Rohrbaugh, and Philip S. Yu. 2024. CORI: CJKV benchmark with Romanization integration - a step towards cross-lingual transfer beyond textual scripts. In *Proceedings of the 2024 Joint International Conference on Computational Linguistics, Language Resources and Evaluation (LREC-COLING 2024)*, pages 4008–4020, Torino, Italia. ELRA and ICCL.

OpenAI. 2024. GPT-4o System Card. `https://openai.com/index/gpt-4o-system-card/`. Accessed: 2025-12-20.

Jonas Pfeiffer, Ivan Vulić, Iryna Gurevych, and Sebastian Ruder. 2021. UNKs everywhere: Adapting multilingual language models to new scripts. In *Proceedings of the 2021 Conference on Empirical Methods in Natural Language Processing*, pages 10186–10203, Online and Punta Cana, Dominican Republic. Association for Computational Linguistics.

Alec Radford, Jong Wook Kim, Chris Hallacy, Aditya Ramesh, Gabriel Goh, Sandhini Agarwal, Girish Sastry, Amanda Askell, Pamela Mishkin, Jack Clark, Gretchen Krueger, and Ilya Sutskever. 2021. Learning transferable visual models from natural language supervision. *Preprint*, arXiv:2103.00020.

Pooyan Rahmanzadehgervi, Logan Bolton, Mohammad Reza Taesiri, and Anh Totti Nguyen. 2025. Vision language models are blind: Failing to translate detailed visual features into words. *Preprint*, arXiv:2407.06581.

Phillip Rust, Jonas Pfeiffer, Ivan Vulić, Sebastian Ruder, and Iryna Gurevych. 2021. How good is your tokenizer? on the monolingual performance of multilingual language models. In *Proceedings of the 59th Annual Meeting of the Association for Computational Linguistics and the 11th International Joint Conference on Natural Language Processing (Volume 1: Long Papers)*, pages 3118–3135, Online. Association for Computational Linguistics.

Fabian David Schmidt, Florian Schneider, Chris Biemann, and Goran Glava. 2025. Mvl-sib: A massively multilingual vision-language benchmark for cross-modal topical matching. *Preprint*, arXiv:2502.12852.

Amanpreet Singh, Vivek Natarajan, Meet Shah, Yu Jiang, Xinlei Chen, Dhruv Batra, Devi Parikh, and Marcus Rohrbach. 2019. Towards vqa models that can read. *Preprint*, arXiv:1904.08920.

Bryan Chen Zhengyu Tan, Zheng Weihua, Zhengyuan Liu, Nancy F. Chen, Hwaran Lee,

Kenny Tsu Wei Choo, and Roy Ka-Wei Lee. 2025. Blend-vis: Benchmarking multimodal cultural understanding in vision language models. *Preprint*, arXiv:2510.11178.

Gemini Team, Petko Georgiev, Ving Ian Lei, Ryan Burnell, Libin Bai, Anmol Gulati, Garrett Tanzer, Damien Vincent, Zhufeng Pan, Shibo Wang, Soroosh Mariooryad, Yifan Ding, Xinyang Geng, Fred Alcober, Roy Frostig, Mark Omernick, Lexi Walker, Cosmin Paduraru, Christina Sorokin, and 1118 others. 2024. Gemini 1.5: Unlocking multimodal understanding across millions of tokens of context.

Ashish V. Thapliyal, Jordi Pont Tuset, Xi Chen, and Radu Soricut. 2022. Crossmodal-3600: A massively multilingual multimodal evaluation dataset. In *Proceedings of the 2022 Conference on Empirical Methods in Natural Language Processing*, pages 715–729, Abu Dhabi, United Arab Emirates. Association for Computational Linguistics.

Ashmal Vayani, Dinura Dissanayake, Hasindri Watawana, Noor Ahsan, Nevasini Sasikumar, Omkar Thawakar, Henok Biadglign Ademtew, Yahya Hmaiti, Amandeep Kumar, Kartik Kuckreja, Mykola Maslych, Wafa Al Ghallabi, Mihail Mihaylov, Chao Qin, Abdelrahman M Shaker, Mike Zhang, Mahardika Krisna Ihsani, Amiel Esplana, Monil Gokani, and 50 others. 2025. All languages matter: Evaluating lmms on culturally diverse 100 languages. *Preprint*, arXiv:2411.16508.

An Vo, Khai-Nguyen Nguyen, Mohammad Reza Taesiri, Vy Tuong Dang, Anh Totti Nguyen, and Daeyoung Kim. 2025. Vision language models are biased. *Preprint*, arXiv:2505.23941.

Ke Wang, Junting Pan, Weikang Shi, Zimu Lu, Mingjie Zhan, and Hongsheng Li. 2024. Measuring multimodal mathematical reasoning with math-vision dataset. *Preprint*, arXiv:2402.14804.

Jason Wei, Xuezhi Wang, Dale Schuurmans, Maarten Bosma, Brian Ichter, Fei Xia, Ed Chi, Quoc Le, and Denny Zhou. 2023. Chain-of-thought prompting elicits reasoning in large language models. *Preprint*, arXiv:2201.11903.

Da Yin, Liunian Harold Li, Ziniu Hu, Nanyun Peng, and Kai-Wei Chang. 2021. Broaden the vision: Geo-diverse visual commonsense reasoning. In *Proceedings of the 2021 Conference on Empirical Methods in Natural Language Processing*, pages 2115–2129, Online and Punta Cana, Dominican Republic. Association for Computational Linguistics.

# A Model Configurations

We provide detailed specifications for all evaluated models to ensure reproducibility. Table 7 summarizes the key hyperparameters used across all experiments.

## A.1 Hardware Infrastructure

**API-Based Models:** All proprietary models were accessed via their respective APIs with no local computational requirements. claude-sonnet-4 was accessed through Google Cloud's Vertex AI service (us-east5 region).
**Local Models:** Evaluated on NVIDIA GH200 Grace Hopper Superchip with:

- **GPU Memory:** 120GB HBM3
- **CUDA Version:** 12.1+
- **Framework:** PyTorch 2.1.0+
- **Memory Management:** `device_map = "cuda"` with `max_memory = {0: "110GB"}`
- **Optimization:** `low_cpu_mem_usage = True`, `trust_remote_code = True`

## A.2 Model-Specific Implementation Details

**Qwen2-VL Series (7B & 72B):**

- **Architecture:** `Qwen2VLForConditionalGeneration`
- **Tokenizer:** `AutoProcessor` with custom Qwen2 tokenizer
- **Image Processing:** Automatic via processor's `apply_chat_template`
- **Input Format:** List-of-dicts conversation structure with image and text content blocks
- **Generation:** Output token trimming (remove input length from generated sequences)

**Llama-3.2-11B-Vision:**

- **Architecture:** `MllamaForConditionalGeneration`
- **Processor:** `AutoProcessor` (handles `MllamaProcessor` automatically)
- **Image Handling:** Passed to processor via `images` parameter
- **Special Tokens:** `<|image|>` tokens inserted automatically by `apply_chat_template`
- **Decoding:** Input tokens trimmed from output: `output_ids[len(input_ids):]`

**LLaVA-OneVision-1.5-8B-Instruct:**

- **Architecture:** `AutoModelForCausalLM` (requires `trust_remote_code=True`)
- **Vision Backbone:** Custom `rice_vit` architecture
- **Processor:** `AutoProcessor` with batch processing
- **Input Format:** Text list + image list with automatic alignment
- **Decoding:** `skip_special_tokens=True`, `clean_up_tokenization_spaces=True`

**InternVL2_5-26B:**

- **Architecture:** `AutoModel` (requires `trust_remote_code=True`)
- **Dynamic Resolution:** Custom preprocessing with aspect ratio preservation
- **Image Preprocessing:** BICUBIC interpolation to 448×448 tiles (min_num=1, max_num=12)
- **Thumbnail Strategy:** Appended for multi-tile images (use_thumbnail=True)
- **Generation:** Native `.chat()` method with pixel_values tensor stack
- **Flash Attention:** Enabled via `use_flash_attn=True`

**Kimi-VL-A3B-Instruct:**

- **Architecture:** `AutoModelForCausalLM`
- **Processor:** Dual tokenizer setup (main tokenizer + processor tokenizer for chat template)
- **Chat Template:** Applied via `processor.tokenizer.apply_chat_template()`

| Model | Organization | Params | Precision | Temp. | Max Tok. | Seed | Inference |
|---|---|---|---|---|---|---|---|
| *Proprietary Frontier Models* | | | | | | | |
| gpt-4o | OpenAI | – | Native | 0.1 | 128 | 42 | API |
| gemini-2.5-flash | Google DeepMind | – | Native | 0.1 | 128 | 42 | Vertex AI |
| claude-sonnet-4 | Anthropic | – | Native | 0.1 | 128 | – | Vertex AI |
| grok-4-1-fast-reasoning | xAI | – | Native | 0.1 | 128 | 42 | API |
| *Open-Weights Models* | | | | | | | |
| Qwen2-VL-7B-Instruct | Alibaba | 7B | bfloat16 | 0.1 | 128 | 42 | Local |
| Qwen2-VL-72B-Instruct | Alibaba | 72B | bfloat16 | 0.1 | 128 | 42 | Local |
| Llama-3.2-11B-Vision | Meta | 11B | bfloat16 | 0.1 | 128 | 42 | Local |
| LLaVA-OneVision-1.5-8B-Instruct | LMMS-Lab | 8B | bfloat16 | 0.1 | 128 | 42 | Local |
| InternVL2_5-26B | OpenGVLab | 26B | bfloat16 | 0.1 | 128 | 42 | Local |
| Kimi-VL-A3B-Instruct | Moonshot AI | 3B | auto | 0.2 | 128 | 42 | Local |

Table 7: Model configuration summary. All models use greedy decoding (temperature $\leq 0.2$). "Local" denotes inference on NVIDIA GH200 with CUDA.

- **Input Structure:** Type-annotated content list: `{"type": "image"}, {"type": "text"}`
- **Temperature:** 0.2
- **Token Management:** `pad_token_id` fallback to `eos_token_id` if unavailable

**API Models (gpt-4o, gemini-2.5-flash, claude-sonnet-4, grok-4-1-fast-reasoning):**

- **Image Encoding:** Base64 JPEG encoding (RGB conversion applied if input is RGBA/other modes)
- **gpt-4o:** OpenAI SDK with `data : image/jpeg;base64` URL format
- **gemini-2.5-flash:** Google Generative AI SDK with `[prompt_text, pil_image]` input list
- **Gemini Safety:** All harm categories set to `BLOCK_NONE` to prevent refusal on research data
- **claude-sonnet-4:** Anthropic Vertex AI client with base64 source in content blocks
- **grok-4-1-fast-reasoning:** OpenAI-compatible SDK (`base_url="https://api.x.ai/v1"`) with image_url format

### A.3 Reproducibility Safeguards

**Random Seeds:** All local models initialize with `seed=42`:

```
torch.manual_seed(42)
torch.cuda.manual_seed_all(42)
```

**API Seeds:** Where supported (gpt-4o, gemini-2.5-flash, grok-4-1-fast-reasoning), `seed=42` parameter passed to API calls.
**Incremental Saving:** Results saved immediately after each prediction via CSV append mode to prevent data loss on system crashes.
**Resume Capability:** All evaluators check for existing CSV files and skip completed instance IDs, enabling seamless resumption after interruptions.
**Deterministic Decoding:** Temperature $\leq$ 0.2 with `do_sample=(temperature > 0)` ensures greedy/near-greedy decoding for reproducibility.

## B Dataset Examples

This section provides representative instances from each of the six task categories in PuMVR. For each example, we present the image stimulus alongside the question and options in all three scripts (Gurmukhi, Shahmukhi, Roman), along with the ground truth answer and human-authored reasoning explanation.

### B.1 Category 1: Visual Analogies

Visual analogies test relational reasoning of the form *A : B :: C : ?*, requiring models to identify the semantic relationship between the first pair and apply it to complete the second pair.

**Example: Category 1 (ID: C1_008)**

**Category:** Visual Analogies

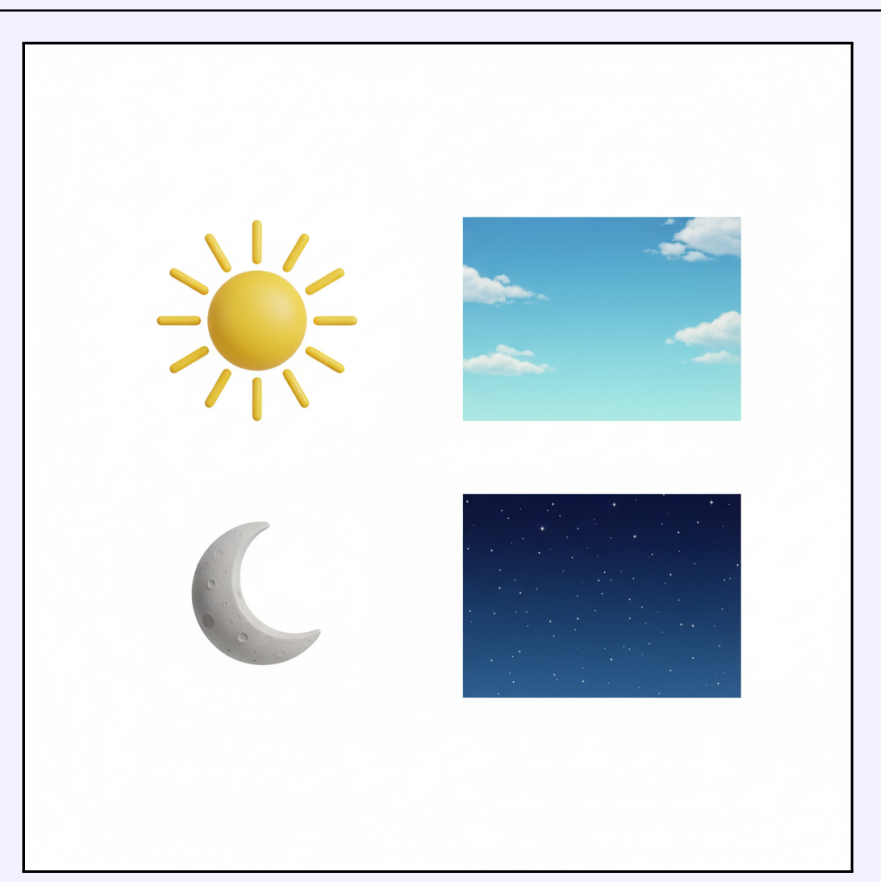

**Gurmukhi:** ਸੂਰਜ : ਦਿਨ :: ਚੰਦ : ?
**A** ਰਾਤ **B** ਪਾਣੀ **C** ਦਿਨ **D** ਮਹੀਨਾ

**Shahmukhi:** سورج : دن :: چنڈ : ؟
**A** رات **B** پانی **C** دن **D** مہینہ

**Roman:** Sooraj : din :: chand : ?
**A** raat **B** paani **C** din **D** mahinaa

**Correct Answer: A**

**Reasoning:** The sun relates to day; the moon relates to night.

### B.2 Category 2: Cultural Object Recognition

This category tests knowledge of Punjabi-specific artifacts, traditional items, and cultural objects that may be unfamiliar to models trained primarily on Western data.

**Example: Category 2 (ID: C2_021)**

**Category:** Cultural Object Recognition

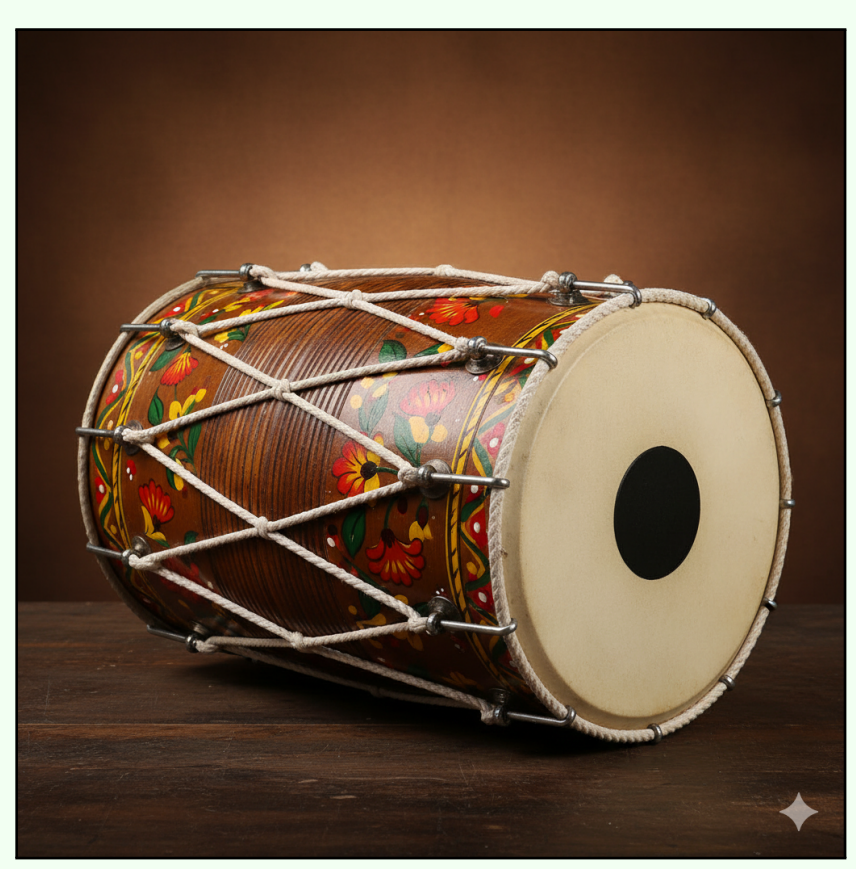

**Gurmukhi:** ਪੰਜਾਬੀ ਸੰਗੀਤ ਵਿੱਚ ਵਰਤੇ ਜਾਣ ਵਾਲੇ ਇਸ ਵੱਡੇ, ਦੋ-ਮੁਖੀ ਸਾਜ਼ ਦਾ ਕੀ ਨਾਮ ਹੈ?
**A** ਤਬਲਾ **B** ਢੋਲ **C** ਢੋਲਕੀ **D** ਮ੍ਰਿਦੰਗਮ

**Shahmukhi:** پنجابی سنگیت وچ ورتے جانے والے اس وڈے، دو-مکھّی ساز دا کی نام ہے؟
**A** تابلا **B** ڈھول **C** ڈھولکی **D** مِردنگم

**Roman:** Punjabi sangeet vich varte jaan vaale is vadde, do-mukhee saaz da kee naam hai?
**A** tablaa **B** dhol
**C** dholkee **D** mridangam

**Correct Answer: B**

**Reasoning:** The 'Dhol' is the iconic large, double-headed drum central to Punjabi music and dance.

## B.3 Category 3: Festival & Celebration Reasoning

This category evaluates cultural knowledge of Punjabi festivals, seasonal celebrations, and associated traditions, requiring both visual recognition and temporal/contextual reasoning.

## B.4 Category 4: Architectural & Landmark Recognition

This category tests geographic and architectural knowledge, requiring models to identify significant Punjabi structures and understand their cultural context.

**Example: Category 3 (ID: C3_001)**

**Category:** Festival & Celebration Reasoning

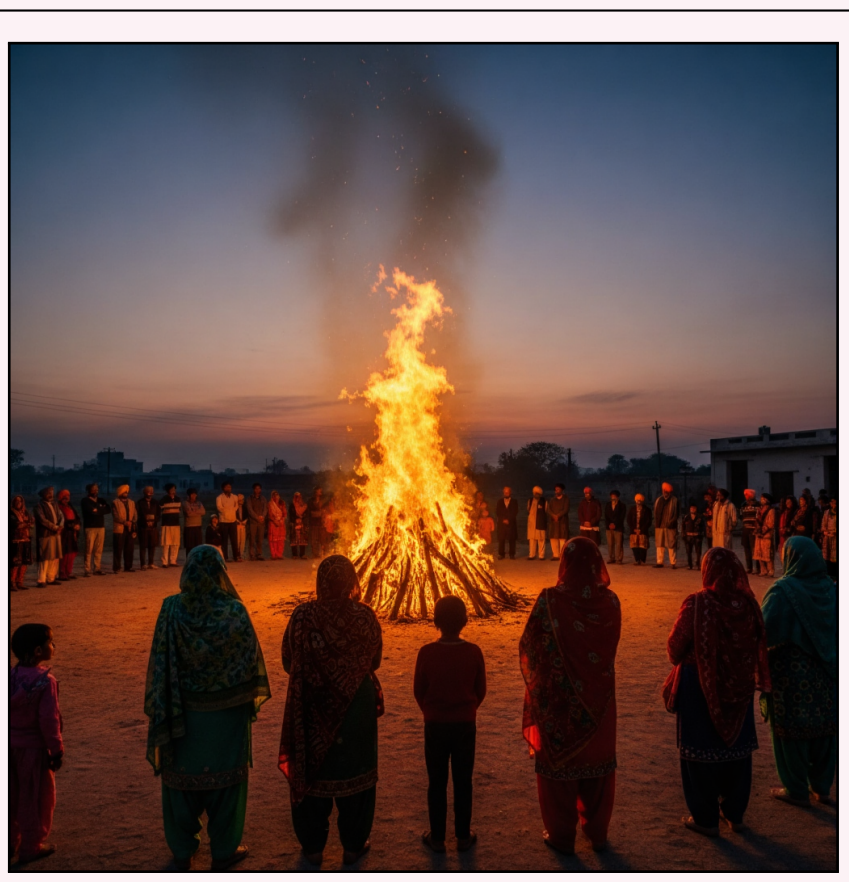

**Gurmukhi:** ਇਸ ਤਸਵੀਰ ਵਿੱਚ ਦਿਖਾਈ ਗਈ ਅੱਗ ਕਿਸ ਤਿਉਹਾਰ ਨਾਲ ਸੰਬੰਧਿਤ ਹੈ?

**A** ਲੋਹੜੀ **B** ਵਿਸਾਖੀ **C** ਦੀਵਾਲੀ **D** ਗੁਰੁਪੁਰਬ

**Shahmukhi:** اس تصویر وچ دکھائی گئی آگ کس تہوار نال متعلق ہے؟

**A** لوہڑی **B** وساکھی **C** دیوالی **D** گروپورب

**Roman:** is tasveer vich dikhaee gaee agg kis tiauhaar naal sanbandhit hai?

**A** lohri **B** visaakhi
**C** deevaalee **D** guroopurb

**Correct Answer: A**

**Reasoning:** Large communal bonfires are the central ritual of Lohri; the image shows people around a large bonfire at dusk.

**Example: Category 4 (ID: C4_001)**

**Category:** Architectural & Landmark Recognition

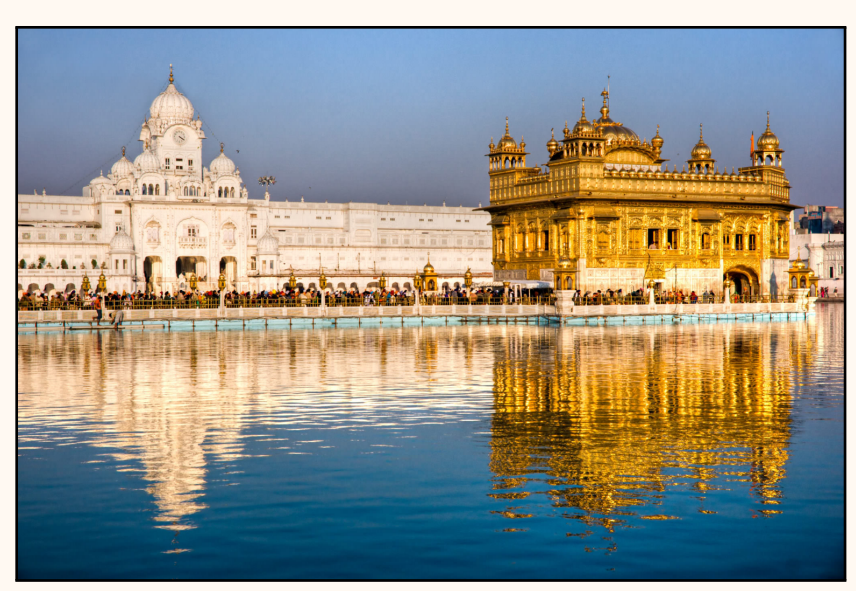

**Gurmukhi:** ਇਸ ਤਸਵੀਰ ਵਿੱਚ ਕਿਹੜਾ ਗੁਰਦੁਆਰਾ ਸਾਹਿਬ ਦਿਖਾਇਆ ਗਿਆ ਹੈ?

**A** ਸ਼੍ਰੀ ਹਰਿਮੰਦਰ ਸਾਹਿਬ
**B** ਸ਼੍ਰੀ ਅਕਾਲ ਤਖ਼ਤ ਸਾਹਿਬ
**C** ਗੁਰਦੁਆਰਾ ਸ਼੍ਰੀ ਬੰਗਲਾ ਸਾਹਿਬ
**D** ਗੁਰਦੁਆਰਾ ਸ਼੍ਰੀ ਨਾਨਕਾਣਾ ਸਾਹਿਬ

**Shahmukhi:** اِس تصویر وِچ کیہڑا گُردوارا صاحِب دِکھایا گیا ہے؟

**A** شری ہرمندر صاحِب
**B** شری اکال تخت صاحِب
**C** گُردوارا شری بنگلا صاحِب
**D** گُردوارا شری ننکانہ صاحِب

**Roman:** is tasveer vichh kehraa gurduaaraa Saahib dikhaai-aa gi-aa hai?

**A** Shree Harimandar Saahib
**B** Shree Akaal Takht Saahib
**C** Gurduaaraa Shree Banglaa Saahib
**D** Gurduaaraa Shree NaanakaaNaa Saahib

**Correct Answer: A**

**Reasoning:** The gold-plated dome and its reflection in the Amrit Sarovar (holy pool) are distinctive identifiers of Harmandir Sahib (Golden Temple).

### B.5 Category 5: Text-in-Image Reasoning

This category specifically tests OCR capabilities and cross-modal integration, where models must read script-specific text embedded in images and reason about it in potentially different query scripts.

### B.6 Category 6: Abstract Visual-Linguistic Reasoning

This category tests basic spatial relationships, logical reasoning, and pattern recognition with Punjabi-language labels, providing a control category for measuring script-specific effects independent of cultural knowledge.

**Example: Category 5 (ID: C5_002)**

**Category:** Text-in-Image Reasoning

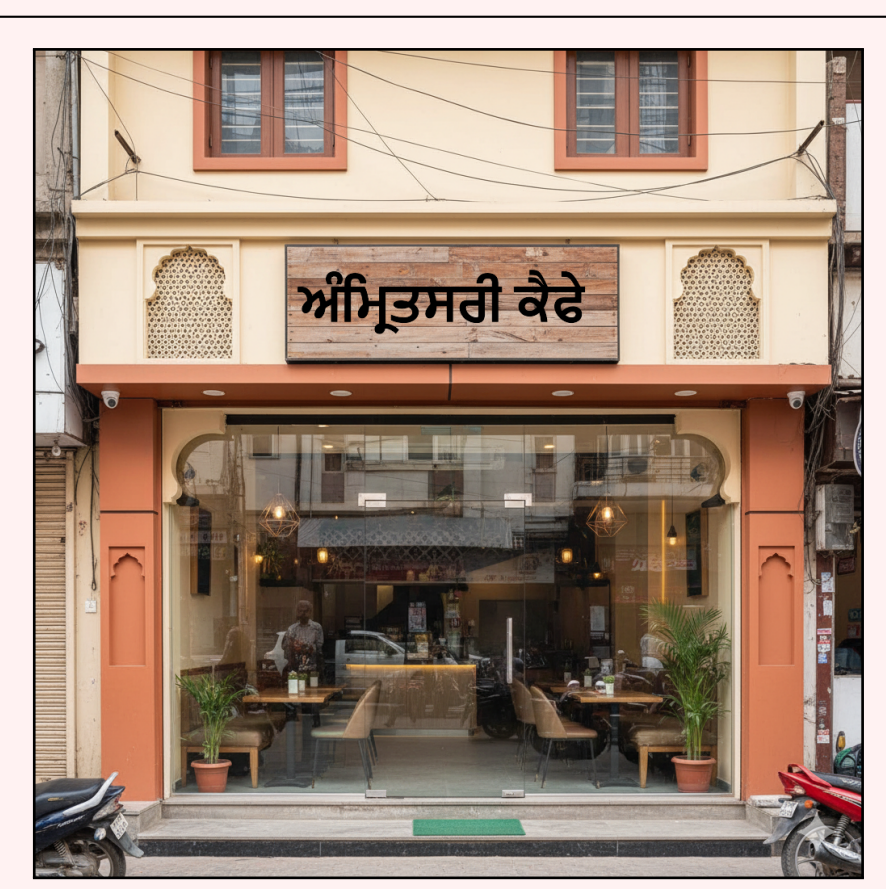

**Gurmukhi:** ਇਸ ਕੈਫੇ ਦਾ ਨਾਮ ਕੀ ਹੈ?
**A** ਲੁਧਿਆਣਾ ਕੈਫੇ **B** ਅੰਮ੍ਰਿਤਸਰੀ ਕੈਫੇ
**C** ਜਲੰਧਰੀ ਕੈਫੇ **D** ਪੰਜਾਬੀ ਕੈਫੇ

**Shahmukhi:** ہے؟ کی نام دا کیفے اس
**A** کیفے لدھیانہ **B** کیفے امرتسری
**C** کیفے جلندھری **D** کیفے پنجابی

**Roman:** is cafe da naam kee hai?
**A** ludhiana cafe **B** amritsari cafe
**C** jalandhari cafe **D** punjabi cafe

**Correct Answer: B**

**Reasoning:** The board displays 'Amritsari Cafe' in Gurmukhi script.

**Example: Category 6 (ID: C6_022)**

**Category:** Abstract Visual-Linguistic Reasoning

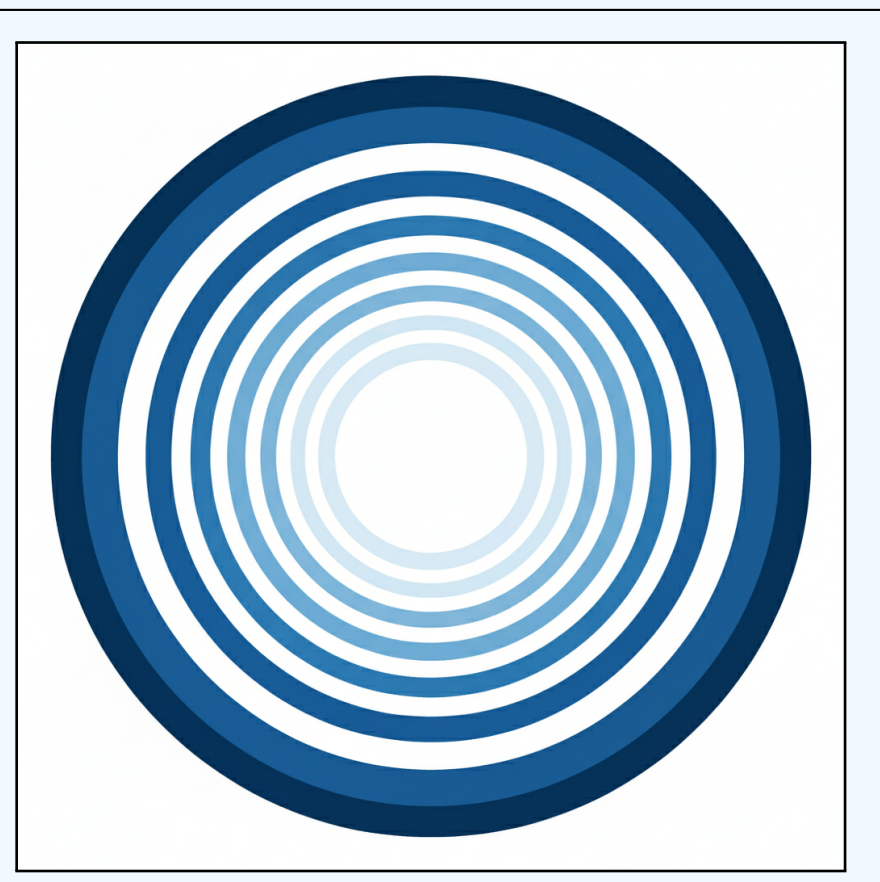

**Gurmukhi:** ਇਸ ਆਕਾਰ ਦੀ ਬਣਤਰ ਕੀ ਹੈ?
**A** ਇੱਕ ਸਿੱਧੀ ਰੇਖਾ **B** ਇੱਕ ਜ਼ਿਗ-ਜ਼ੈਗ ਪੈਟਰਨ
**C** ਇੱਕ ਸਪਾਈਰਲ **D** ਇੱਕ ਸਧਾਰਨ ਚੱਕਰ

**Shahmukhi:** ہے؟ کی بنتر دی آکار اس
**A** ریکھا سدّھی اکّ **B** پیٹرن زگ-زیگ اکّ
**C** سپائیرل اکّ **D** چکر سدھارن اکّ

**Roman:** is aakaar di banatar kee hai?
**A** ikk siddhi rekha **B** ikk zig-zag pattern
**C** ikk spiral **D** ikk sadharan chakkar

**Correct Answer: C**

**Reasoning:** A spiral pattern.

## C Complete Results Tables

### C.1 Experiment 1: Per-Script Accuracy Details

Table 8 presents the complete accuracy results across all three scripts for each model.

| Model | Gur. | Shah. | Rom. | SCR |
|---|---|---|---|---|
| gpt-4o | 90.93 | 86.67 | 86.93 | 78.13 |
| gemini-2.5-flash | 93.87 | 88.53 | 92.27 | 85.60 |
| claude-sonnet-4 | 91.47 | 84.76 | 85.83 | 77.09 |
| grok-4-1-fast-reasoning | 92.00 | 84.53 | 90.13 | 79.73 |
| Qwen2-VL-7B-Instruct | 68.00 | 58.40 | 65.07 | 41.33 |
| Llama-3.2-11B-Vision | 62.13 | 45.87 | 60.27 | 27.47 |
| LLaVA-OneVision-1.5-8B-Instruct | 71.20 | 63.47 | 61.07 | 41.33 |
| InternVL2_5-26B | 58.93 | 45.60 | 59.73 | 28.00 |
| Kimi-VL-A3B-Instruct | 51.20 | 48.53 | 59.73 | 24.80 |
| Qwen2-VL-72B-Instruct | 83.54 | 79.43 | 80.25 | 54.67 |
| **Average** | **76.32** | **68.58** | **74.13** | **53.82** |

Table 8: Zero-shot accuracy (%) across scripts. SCR = Script Consistency Rate.

### C.2 Experiment 3: Transfer Efficiency Matrix

Table 10 shows Transfer Efficiency (TE) values for all few-shot transfer conditions.

### C.3 Experiment 4: Script Disparity Index Results

Table 11 shows accuracy across all 9 image-script and question-script combinations for Category 5 (Text-in-Image Reasoning).

## D Full Experimental Prompts

This section provides the exact prompt templates used across all five experiments. To ensure high-fidelity evaluation, we enforced strict formatting constraints.

### D.1 Notation and Placeholder Definitions

To maintain clarity across the templates provided in this appendix, we use the following placeholders to denote dynamically injected content:

- `{question}`: The Punjabi query translated into the specific target script (Gurmukhi, Shahmukhi, or Roman).
- `{formatted_options}`: A list of four multiple-choice candidates. Each candidate is presented on a new line without alphabetical prefixes (A, B, C, D) or numerical markers.
- `{script_name}`: The English name of the target script (e.g., "Gurmukhi", "Shahmukhi", or "Roman").
- `{Script Instruction}`: A script-specific directive used in Experiments to trigger internal script-specific reasoning. The values used were:
  - **Gurmukhi:** "ਗੁਰਮੁਖੀ ਵਿੱਚ"
  - **Shahmukhi:** "وچ مکھی شاه"
  - **Roman:** "in Roman script"

### D.2 Design Rationale: Text-Based vs. Letter-Based Selection

In all experiments, models were instructed to output the *exact text* of the correct option rather than a single identifier (e.g., "A", "B"). This design choice was made for two primary reasons:

1. **Script Comprehension Verification:** A model might correctly guess a letter (25% probability) without truly processing the script. Forcing the model to reproduce the script-specific text ensures it can parse and generate the target orthography.
2. **Failure Mode Analysis:** By requiring text output, we could identify cases where models produced "hallucinated" characters or mixed scripts, data that would be lost if restricted to single-letter outputs.

### D.3 Experiment 1: Baseline Script Gap

The primary benchmark used the following template with English instructions and script-specific constraints.

**Exp 1: Standard Prompt**

```
Question: {question}

Options:
{formatted_options}

Task: Identify the correct option based on the image.

Constraint: {Answer in [Script Name] script.} Copy the exact text of the correct option. Do not explain.

CRITICAL RULES:
1. Answer MUST be in the SAME script as the question
2. Copy EXACTLY one option from above - character by character
3. NO explanations, NO extra words, NO English translations
4. NO letters like A), B), C) or numbers
5. Output ONLY the option text, nothing else

Your answer (copy exact text from options):
Answer:
```

### D.4 Experiment 2: Native Instruction Prompting

This experiment used similar prompts as experiment 1, without the image input to test model performance and establish a baseline.

**Exp 2: Native Script Instructions**

```
Question {Script Instruction}: {question}

Options:
{formatted_options}

CRITICAL RULES:
1. Answer MUST be in the SAME script as the question
2. Copy EXACTLY one option from above - character by character
3. NO explanations, NO extra words, NO English translations
4. NO letters like A), B), C) or numbers
5. Output ONLY the option text, nothing else

Your answer (copy exact text from options):
```

### D.5 Experiment 3: System Prompting (Few-Shot)

For experiments involving system-level instructions, the following persona-based prompt was utilized.

**Exp 3: System Prompt**

```
You are a precise answering assistant.
You will be given a visual question and options.
You must output ONLY the exact text of the correct option in {Script Name} script.

CRITICAL RULES:
1. NO reasoning.
2. NO explanations.
3. NO option letters (like A, B).
4. Output strictly the option text.
```

### D.6 Experiment 4: OCR-Aware Transcription

This prompt forced the model to first "look and write" before deciding, to decouple visual recognition from reasoning.

**Exp 4: OCR + Decision Prompt**

```
Question {Script Instruction}: {question}

Options: {formatted_options}

TASK INSTRUCTIONS:
1. LOOK at the image text carefully.
2. TRANSCRIPTION: Write down exactly what text you see in the image (if any).
3. DECISION: Choose the correct option from the list above.

REQUIRED OUTPUT FORMAT:
OCR: [Write the text from the image here]
1. You are taking a multiple choice test.
2. Select the Correct Option from the list above.
3. Your output must be IDENTICAL to one of the options.
4. DO NOT provide reasoning. DO NOT write 'Answer:'.
5. DO NOT use A), B), C), D).

ANSWER: [Write ONLY the selected option text here]
```

## D.7 Experiment 5: Chain-of-Thought (CoT)

Models were prompted to follow a 5-step reasoning process. The instructions were provided in the same script as the question.

**Exp 5: Gurmukhi CoT Template**

ਤੁਹਾਨੂੰ ਇਹਨਾਂ ਕਦਮਾਂ ਦੀ ਸਪੱਸ਼ਟ ਤੌਰ ’ਤੇ ਪਾਲਣਾ ਕਰਕੇ ਇਸ ਮਲਟੀਮੋਡਲ ਸਮੱਸਿਆ ਨੂੰ ਹੱਲ ਕਰਨਾ ਹੈ:

1. ਵਿਜ਼ੂਅਲ ਗਰਾਊਂਡਿੰਗ: ਚਿੱਤਰ ਦੇ ਆਧਾਰ ’ਤੇ, 3 ਸਭ ਤੋਂ ਪ੍ਰਮੁੱਖ ਵਸਤੂਆਂ, ਦ੍ਰਿਸ਼ਾਂ ਜਾਂ ਕਿਰਿਆਵਾਂ ਦੀ ਸੂਚੀ ਬਣਾਓ।
2. ਭਾਸ਼ਾਈ ਪਾਰਸਿੰਗ: ਇਹ ਯਕੀਨੀ ਬਣਾਉਣ ਲਈ ਕਿ ਤੁਸੀਂ ਇਸਨੂੰ ਸਮਝਦੇ ਹੋ, ਆਪਣੇ ਸ਼ਬਦਾਂ ਵਿੱਚ ਪ੍ਰਸ਼ਨ ਨੂੰ ਦੁਬਾਰਾ ਲਿਖੋ।
3. ਸੰਬੰਧ ਐਬਸਟਰੈਕਸ਼ਨ: ਪ੍ਰਸ਼ਨ ਦੁਆਰਾ ਪਰਖਣ ਵਾਲੇ ਮੁੱਖ ਸੰਬੰਧ ਜਾਂ ਸੰਕਲਪ ਦੀ ਪਛਾਣ ਕਰੋ।
4. ਕਰਾਸ-ਮਾਡਲ ਏਕੀਕਰਣ: ਸਮਝਾਓ ਕਿ ਕਦਮ 1 ਦੇ ਵਿਜ਼ੂਅਲ ਤੱਤ ਤੁਹਾਨੂੰ ਕਦਮ 2 ਦੇ ਪ੍ਰਸ਼ਨ ਦਾ ਉੱਤਰ ਦੇਣ ਵਿੱਚ ਕਿਵੇਂ ਮਦਦ ਕਰਦੇ ਹਨ।
5. ਕਟੌਤੀ: ਕਦਮ 1-4 ਦੇ ਆਧਾਰ ’ਤੇ, ਉੱਤਰ ਕੱਢੋ।

ਅੰਤਿਮ ਉੱਤਰ: ਅੰਤਿਮ ਵਿਕਲਪ ਟੈਕਸਟ ਦੱਸੋ ( ਦਿੱਤੇ ਗਏ ਵਿਕਲਪਾਂ ਤੋਂ ਬਿਲਕੁਲ ਕਾਪੀ ਕਰੋ )।

ਹੁਣ, ਹੇਠਾਂ ਦਿੱਤੀ ਸਮੱਸਿਆ ਨੂੰ ਕਦਮ-ਦਰ-ਕਦਮ ਹੱਲ ਕਰੋ।

**Exp 5: Shahmukhi CoT Template**

پالنا تے’ طور سپشٹ دی قدماں ایہناں توہانوں
ہے کرنا حل نوں سمسیا ماڈل ملٹی اِس کرکے:

1. تے’ آدھار دے چِتر :گراؤنڈنگ وِژوئل
کریاواں جاں درشاں وستواں، پرمکھ توں سبھ 3
بناؤ۔ سوچی دی
2. کہ لئی بناؤن یقینی ایہہ :پارسنگ بھاشائی
پرشن وِچ شبداں اپنے ہو، سمجھدے اِسنوں تسیں
لِکھو۔ دوبارہ نوں
3. پرکھن دوارا پرشن :ایبسٹریکشن سنبندھ
کرو۔ پچھان دی سنکلپ جاں سنبندھ مُکھ والے
4. 1 قدم کہ سمجھاؤ :ایکیکرن کراس-ماڈل
اُتر دا پرشن دے 2 قدم توہانوں تتّ وِژوئل دے
ہن۔ کردے مدد کِویں وِچ دین
5. کڈھو۔ اُتر تے’ آدھار دے 1-4 قدم :کٹوتی

دِتّے ) دسّو ٹیکسٹ وکلپ انتم :اُتر انتم
کرو کاپی بِلکُل توں وکلپاں گئے (۔

قدم-در-قدم نوں سمسیا دِتّی ہیٹھاں ہُن،
کرو۔ حل

**Exp 5: Roman CoT Template**

tuhannu ehnaan kadmaan dee spasht taur ’te paalna karke is malteemodal samassya nu hall karna hai:

1. vizooal graaoonding: chittar de aadhaar ’te, 3 sabh ton pramukkh vastuaan, drishaan jaan kiriavaan dee soochee banao.
2. bhashai paarsing: eh yakeenee banaun laee ki tuseen isnu samajhde ho, aapne shabdaan vich prashan nu dubaara likho.
3. sambandh aibstraikshan: prashan duara parkhan vaale mukkh sambandh jaan sankalp dee pachhaan karo.
4. kraas-modal ekeekaran: samjhao ki kadam 1 de vizooal tatt tuhannu kadam 2 de prashan da uttar den vich kiven madad karde han.
5. katauti: kadam 1-4 de aadhaar ’te, uttar kaddho.

antim uttar: antim vikalp taikst dasso (ditte gae vikalpaan ton bilkul kaapee karo).

hun, hethaan dittee samassya nu kadam-dar-kadam hall karo.

# E Statistical Significance Testing

Table 9 reports McNemar’s test results for all pairwise script comparisons across all 10 models. We apply McNemar’s test treating per-instance correctness as paired binary observations across 375 instances. For each model, we test three script pairs: Gurmukhi vs. Shahmukhi, Gurmukhi vs. Roman, and Shahmukhi vs. Roman. The chi-squared variant is used when discordant pairs exceed 25; the exact variant otherwise.

Several patterns emerge. First, the Gurmukhi–Shahmukhi gap is the most consistently significant pair across models, confirming Shahmukhi as the most undertrained script. Second, the Gurmukhi–Roman gap is significant for some models (claude-sonnet-4, LLaVA) but not others, suggesting variable Roman script coverage across model fam-ilies. Third, Qwen2-VL-72B-Instruct is the only model where no pair reaches significance, consistent with its more balanced cross-script performance observed in Table 8. These results validate that the Script Gap reported in Section 5.1 is statistically robust and not an artifact of the dataset size.

| **Model** | **Script Pair** | ***p*-value** | **Sig.** |
|---|---|---|---|
| gpt-4o | Gur vs Shah | 0.025 | * |
| gpt-4o | Gur vs Rom | 0.041 | * |
| gpt-4o | Shah vs Rom | 0.883 | ns |
| gemini-2.5-flash | Gur vs Shah | 0.001 | ** |
| gemini-2.5-flash | Gur vs Rom | 0.146 | ns |
| gemini-2.5-flash | Shah vs Rom | 0.038 | * |
| claude-sonnet-4 | Gur vs Shah | <0.001 | *** |
| claude-sonnet-4 | Gur vs Rom | 0.003 | ** |
| claude-sonnet-4 | Shah vs Rom | 0.892 | ns |
| grok-4-1-fast-reasoning | Gur vs Shah | <0.001 | *** |
| grok-4-1-fast-reasoning | Gur vs Rom | 0.265 | ns |
| grok-4-1-fast-reasoning | Shah vs Rom | 0.006 | ** |
| Qwen2-VL-7B-Instruct | Gur vs Shah | <0.001 | *** |
| Qwen2-VL-7B-Instruct | Gur vs Rom | 0.343 | ns |
| Qwen2-VL-7B-Instruct | Shah vs Rom | 0.044 | * |
| Llama-3.2-11B-Vision | Gur vs Shah | <0.001 | *** |
| Llama-3.2-11B-Vision | Gur vs Rom | 0.608 | ns |
| Llama-3.2-11B-Vision | Shah vs Rom | <0.001 | *** |
| LLaVA-OneVision-1.5-8B-Instruct | Gur vs Shah | 0.004 | ** |
| LLaVA-OneVision-1.5-8B-Instruct | Gur vs Rom | <0.001 | *** |
| LLaVA-OneVision-1.5-8B-Instruct | Shah vs Rom | 0.415 | ns |
| InternVL2_5-26B | Gur vs Shah | <0.001 | *** |
| InternVL2_5-26B | Gur vs Rom | 0.857 | ns |
| InternVL2_5-26B | Shah vs Rom | <0.001 | *** |
| Kimi-VL-A3B-Instruct | Gur vs Shah | 0.430 | ns |
| Kimi-VL-A3B-Instruct | Gur vs Rom | 0.006 | ** |
| Kimi-VL-A3B-Instruct | Shah vs Rom | 0.001 | ** |
| Qwen2-VL-72B-Instruct | Gur vs Shah | 0.131 | ns |
| Qwen2-VL-72B-Instruct | Gur vs Rom | 0.208 | ns |
| Qwen2-VL-72B-Instruct | Shah vs Rom | 0.892 | ns |

Table 9: McNemar's test across all script pairs and models. *** $p<0.001$, ** $p<0.01$, * $p<0.05$, ns = not significant.

| Model | S→G | R→G | G→S | R→S | G→R | S→R | Avg. TE |
|---|---|---|---|---|---|---|---|
| gpt-4o | 100.85 | 101.70 | 100.88 | 100.29 | 101.45 | 100.87 | **101.01** |
| gemini-2.5-flash | 98.04 | 98.88 | 99.12 | 100.00 | 99.15 | 100.00 | 99.20 |
| claude-sonnet-4 | 96.80 | 99.42 | 101.25 | 101.56 | 99.71 | 100.54 | 99.88 |
| grok-4-1-fast-reasoning | 100.29 | 100.87 | 100.31 | 99.38 | 89.52 | 95.81 | 97.70 |
| Qwen2-VL-7B-Instruct | 97.56 | 97.56 | 102.76 | 96.77 | 100.00 | 98.35 | 98.83 |
| Llama-3.2-11B-Vision | 101.86 | 101.86 | 67.22 | 97.51 | 87.85 | 100.00 | 92.72 |
| LLaVA-OneVision-1.5-8B-Instruct | 103.07 | 100.38 | 102.06 | 97.12 | 100.00 | 100.85 | 100.58 |
| InternVL2_5-26B | 96.51 | 100.58 | 102.76 | 100.00 | 100.00 | 100.00 | 99.98 |
| Kimi-VL-A3B-Instruct | 107.87 | 110.11 | 101.70 | 105.68 | 76.23 | 88.34 | 98.32 |
| Qwen2-VL-72B-Instruct | 154.76 | 153.17 | 81.20 | 97.37 | 88.70 | 100.33 | 112.59 |
| **Average** | **105.76** | **106.45** | **95.93** | **99.57** | **94.26** | **98.51** | **100.08** |

Table 10: Transfer Efficiency (%) across script pairs. Notation: G=Gurmukhi, S=Shahmukhi, R=Roman.

| Model | Image Script | Gur. Q | Shah. Q | Rom. Q | SDI |
|---|---|---|---|---|---|
| gpt-4o | Gurmukhi | 100.00 | 100.00 | 95.00 | 5.00 |
| | Shahmukhi | 75.00 | 90.00 | 95.00 | 20.00 |
| | Roman | 100.00 | 100.00 | 100.00 | 0.00 |
| gemini-2.5-flash | Gurmukhi | 100.00 | 100.00 | 100.00 | 0.00 |
| | Shahmukhi | 85.00 | 85.00 | 85.00 | 0.00 |
| | Roman | 100.00 | 95.00 | 100.00 | 5.00 |
| claude-sonnet-4 | Gurmukhi | 100.00 | 95.00 | 100.00 | 5.00 |
| | Shahmukhi | 70.00 | 80.00 | 75.00 | 10.00 |
| | Roman | 100.00 | 100.00 | 100.00 | 0.00 |
| grok-4-1-fast-reasoning | Gurmukhi | 90.00 | 70.00 | 90.00 | 20.00 |
| | Shahmukhi | 75.00 | 75.00 | 60.00 | 15.00 |
| | Roman | 100.00 | 100.00 | 100.00 | 0.00 |
| Qwen2-VL-7B-Instruct | Gurmukhi | 85.00 | 35.00 | 85.00 | 50.00 |
| | Shahmukhi | 70.00 | 65.00 | 75.00 | 10.00 |
| | Roman | 60.00 | 60.00 | 90.00 | 30.00 |
| Llama-3.2-11B-Vision | Gurmukhi | 90.00 | 40.00 | 95.00 | 55.00 |
| | Shahmukhi | 70.00 | 70.00 | 60.00 | 10.00 |
| | Roman | 75.00 | 45.00 | 90.00 | 45.00 |
| LLaVA-OneVision-1.5-8B-Instruct | Gurmukhi | 70.00 | 35.00 | 75.00 | 40.00 |
| | Shahmukhi | 45.00 | 55.00 | 50.00 | 10.00 |
| | Roman | 100.00 | 90.00 | 95.00 | 10.00 |
| InternVL2_5-26B | Gurmukhi | 80.00 | 55.00 | 80.00 | 25.00 |
| | Shahmukhi | 25.00 | 40.00 | 60.00 | 35.00 |
| | Roman | 95.00 | 100.00 | 90.00 | 10.00 |
| Kimi-VL-A3B-Instruct | Gurmukhi | 70.00 | 40.00 | 75.00 | 35.00 |
| | Shahmukhi | 80.00 | 40.00 | 60.00 | 40.00 |
| | Roman | 95.00 | 60.00 | 75.00 | 35.00 |
| Qwen2-VL-72B-Instruct | Gurmukhi | 50.00 | 30.00 | 75.00 | 45.00 |
| | Shahmukhi | 85.00 | 65.00 | 85.00 | 20.00 |
| | Roman | 100.00 | 90.00 | 95.00 | 10.00 |

Table 11: OCR script mismatch performance (%). Each row shows accuracy when the image contains text in the specified script, while the question is posed in Gurmukhi (Gur. Q), Shahmukhi (Shah. Q), or Roman (Rom. Q). SDI = Script Disparity Index (max - min accuracy for that image script).